\documentclass{article}

\usepackage{microtype}
\usepackage{graphicx}
\usepackage{subfigure}
\usepackage{booktabs} % for professional tables
\usepackage{graphbox} % vertical alignment
\usepackage{float} % force position

\usepackage{hyperref}

\usepackage[accepted]{mlsys2023}

\mlsystitlerunning{Towards Modular Machine Learning Solution Development: Benefits and Trade-offs}

\begin{document}

\twocolumn[
\mlsystitle{Towards Modular Machine Learning Solution Development: Benefits and Trade-offs}

% It is OKAY to include author information, even for blind
% submissions: the style file will automatically remove it for you
% unless you've provided the [accepted] option to the mlsys2023
% package.

% List of affiliations: The first argument should be a (short)
% identifier you will use later to specify author affiliations
% Academic affiliations should list Department, University, City, Region, Country
% Industry affiliations should list Company, City, Region, Country

% You can specify symbols, otherwise they are numbered in order.
% Ideally, you should not use this facility. Affiliations will be numbered
% in order of appearance and this is the preferred way.
\mlsyssetsymbol{equal}{*}

\begin{mlsysauthorlist}
\mlsysauthor{Samiyuru Menik\footnotemark[1]}{}
\mlsysauthor{Lakshmish Ramaswamy\footnotemark[1]}{}
\end{mlsysauthorlist}

% \mlsysaffiliation{uga}{School of Computing, University of Georgia, Athens, GA, USA}

\mlsyscorrespondingauthor{Samiyuru Menik}{sm19812@uga.edu}
% \mlsyscorrespondingauthor{Lakshmish Ramaswamy}{laksmr@uga.edu}

% You may provide any keywords that you
% find helpful for describing your paper; these are used to populate
% the "keywords" metadata in the PDF but will not be shown in the document
\mlsyskeywords{machine learning, modularity, machine learning solution engineering, machine learning systems}

\vskip 0.3in

\begin{abstract}
% This is to be rewritten after the paper is complete.
% I am writing an initial abstract to get a sense of the structure and layout a vision.
Machine learning technologies have demonstrated immense capabilities in various domains. They play a key role in the success of modern businesses. However, adoption of machine learning technologies has a lot of untouched potential. Cost of developing custom machine learning solutions that solve unique business problems is a major inhibitor to far-reaching adoption of machine learning technologies. We recognize that the monolithic nature prevalent in today's machine learning applications stands in the way of efficient and cost effective customized machine learning solution development. In this work we explore the benefits of modular machine learning solutions and discuss how modular machine learning solutions can overcome some of the major solution engineering limitations of monolithic machine learning solutions. We analyze the trade-offs between modular and monolithic machine learning solutions through three deep learning problems; one text based and the two image based. Our experimental results show that modular machine learning solutions have a promising potential to reap the solution engineering advantages of modularity while gaining performance and data advantages in a way the monolithic machine learning solutions do not permit.
\end{abstract}
]

% this must go after the closing bracket ] following \twocolumn[ ...

% This command actually creates the footnote in the first column
% listing the affiliations and the copyright notice.
% The command takes one argument, which is text to display at the start of the footnote.
% The \mlsysEqualContribution command is standard text for equal contribution.
% Remove it (just {}) if you do not need this facility.

%\printAffiliationsAndNotice{}  % leave blank if no need to mention equal contribution
% \printAffiliationsAndNotice{} % otherwise use the standard text.

\footnotetext[1]{School of Computing, University of Georgia,
Athens, Georgia, United States. Correspondence to: Samiyuru Menik $<$sm19812@uga.edu$>$}

%%%%%%%%%%%%%%%%%%%%%%%%%%%%%%%%%%%%%%%%%%%%%%%%%%%
%%%%%%%%%%%%%%%%%%%%%%%%%%%%%%%%%%%%%%%%%%%%%%%%%%%
%% Beginning of content.
%%%%%%%%%%%%%%%%%%%%%%%%%%%%%%%%%%%%%%%%%%%%%%%%%%%
%%%%%%%%%%%%%%%%%%%%%%%%%%%%%%%%%%%%%%%%%%%%%%%%%%%

%%%%%%%%%%%%%%%%%%%%%%%%%%%%%%%%%%%%%%%%%%%%%%%%%%%%%%%
%%%%%%%%%%%%%%%%%%%%%%%%%%%%%%%%%%%%%%%%%%%%%%%%%%%%%%%
%%%%%%%%%%%%%%%%%%%%%%%%%%%%%%%%%%%%%%%%%%%%%%%%%%%%%%%
%%%%%%%%%%%%%%%%%%%%%%%%%%%%%%%%%%%%%%%%%%%%%%%%%%%%%%%
%%%%%%%%%%%%%%%%%%%%%%%%%%%%%%%%%%%%%%%%%%%%%%%%%%%%%%%
%%%%%%%%%%%%%%%%%%%%%%%%%%%%%%%%%%%%%%%%%%%%%%%%%%%%%%%
%%%%%%%%%%%%%%%%%%%%%%%%%%%%%%%%%%%%%%%%%%%%%%%%%%%%%%%

\section{Introduction}

% Todo: Say that we are proposing distillation to overcome some of the downsides of modular solutions.

% Machine learning is widely applicable. In many industries.
Machine learning (ML) has gained a lot of attention over the past years. Machine learning technologies have become a part of many organizational workflows and day to day tasks of individuals knowingly or unknowingly. Big tech companies and academic entities are taking the lead in developing cutting edge ML technologies that push the boundaries of what ML can accomplish.
% You can give more examples and citations:
Cutting edge computer vision and language modeling technologies provide a good example for this.
% There is a lot of untouched potential. Cost of ML solution development is a key part of the issue.
% Argument:> ML is very applicable, Big tech push boundary, Untouched potential, where is this potential?
% Argument: Attention, applied, big tech push|, untouch potential in app domains, adoption,
Beyond the heightened attention, existing applications and large scale organization and academic driven developments, there is a lot of untouched potential to ML. We believe that this potential lies in ground level application domains that are usually away from the mainstream attention.
% What are these domains?
Think of organizations that operate in non-tech focused business domains. These can include educational and research institutes, healthcare facilities, transportation units and government organizations. These organizations have a number of workflows that can be improved using ML technologies. Depending on the scale, these organizations are more likely to have a traditional information technology and software engineering staff to fulfill the tech requirements. However, these organizations may not have the budgets, resources and expertise to focus on producing custom ML solutions for their workflows.
% This maybe is a good place to paste content from the motivation.
% Make ML accessible to a wider audience. What does it enable? How to enable it? 
Making cutting edge technologies accessible to a wider range of organizations and different organizational levels and individuals that operate in a wide range of domains is one of the big challenges that ML as a field face today. Enabling wide scale adoption of ML technologies has the potential of achieving the next level of business process optimizations, service quality improvements and user experiences. As an example, today's high level decision makers of large organizations already use machine learning technologies intensively in their decision making processes. But how much of these technologies are practical and cost efficient to be implemented for the use of lower levels of the organizational hierarchy? As another more concrete example, a large scale organization may implement a cutting edge machine learning solution to filter resumes in their hiring process. How practical is it to access such technologies to implement a similar system for a smaller scale organization that matches their customized business requirements? Recurring theme here is that, when making machine learning technologies more accessible, 1/ cutting edge technologies should be accessible to a wider audience through means that are easy to grasp 2/ it should be possible to implement customized machine learning solutions that match business requirements at a lower cost.
% Improving engineering practices is key to solving issues.
We believe that further improving software engineering practices in machine learning solution development, especially focusing on deep learning approaches, can provide an effective solution to the aforementioned challenge.
% Today's machine learning is monolithic. State of the art pushes towards monolithic solutions with end to end machine learning ideas.
Today's prevalent deep learning solutions are monolithic. These solutions are mostly large black boxes that produce the right answer to a given problem when fed with a large amount of relevant data.
% From the modularity section:
% State of the art in ML tends towards learning ML solutions end to end. These technologies enable creating large parameterized models that can be trained with large amounts of data to produce impressive results.
% End to end learning ideas and examples.
Major influence for this trend is coming from end to end deep learning technologies. Unlike traditional ML methods that involve time consuming and labor intensive feature engineering steps, end to end machine learning attempts use a larger single model that aims to learn all intermediate steps required to solve the problem at hand without needing much hand engineering during the learning process. This is usually achieved by employing large parameterized models that can take advantage of large amounts of training data. This way of developing ML solutions significantly cut down the human effort needed to implement ML solutions that produce impressive results for a variety of ML problems. However, this approach of developing ML solutions has a set of key disadvantages as well. They became especially apparent when developing ML solutions outside of cutting edge tech focused organizations. 

\begin{figure*}[ht]
    \centering
    \includegraphics[width=0.90\textwidth]{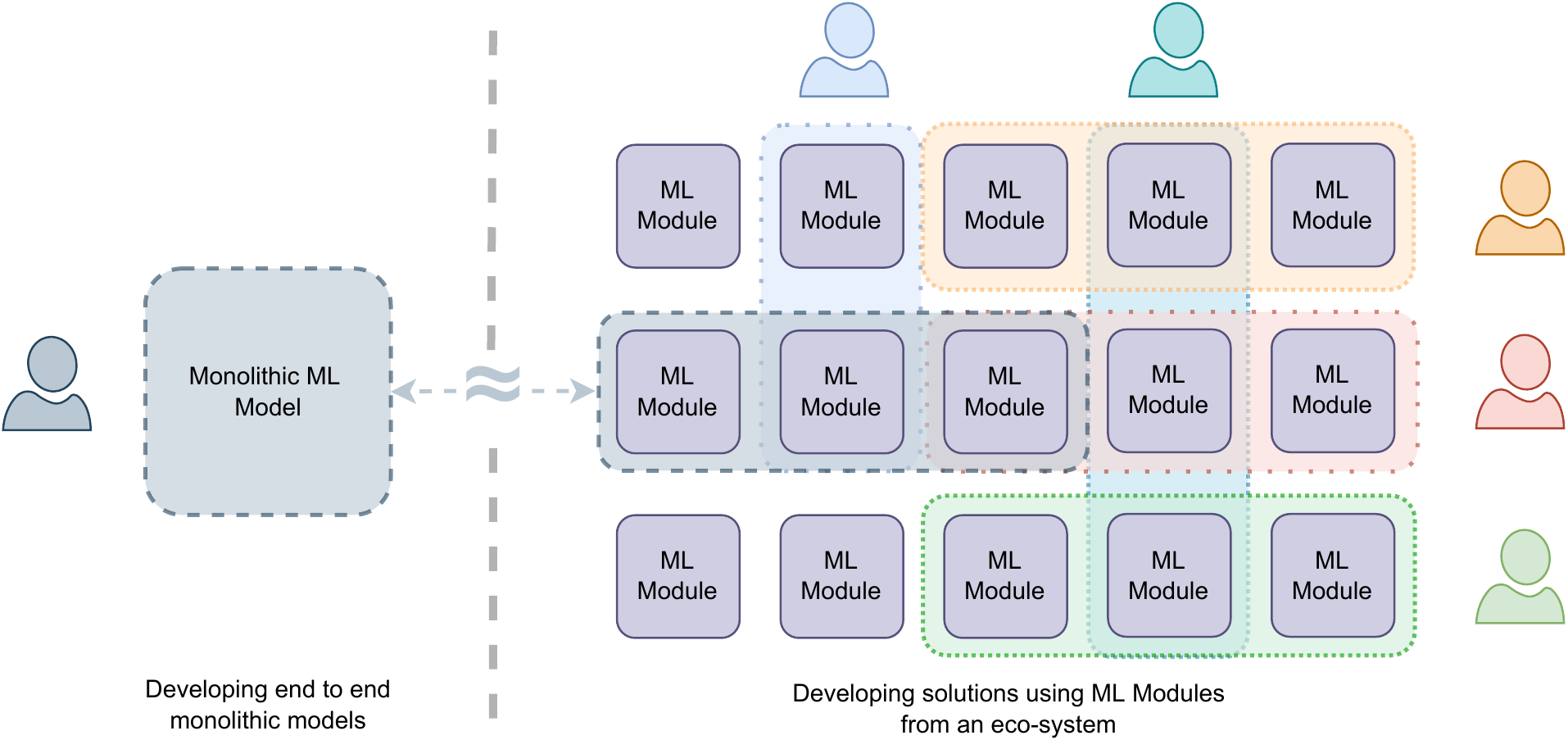}
    \caption{Visualization of monolithic ML solution development in comparison to modular ML solution development in an echo-system. In the modular solution development paradigm, ML modules are reused to develop different solutions by different users enabling enhanced accessibility to technologies.}
    \label{fig:modularity_banner_image}
\end{figure*}

% Give examples to the worries.
One of the key limitations of prevalent end to end machine learning solutions is the lack of modularity. The lack of modularity results in a number of other critical engineering challenges. Usually one monolithic solution is only applicable to a single problem. Since the solution can not be broken down into semantically meaningful and more general submodules, the effort that went into the system can not be easily reused in other contexts. Also, it is not possible to replace parts of the system as new technologies are introduced with improved performance characteristics. Further, the solutions have to rely on specific datasets that are focused on the specific problem.
% Engineering always recommends modularity. There are a number of reasons for choosing modularity. These advantages can be incorporated in ML solutions as well.
Traditionally engineering practices encourage modular solutions in general to overcome the aforementioned challenges and develop more maintainable and cost efficient solutions. We believe that encouraging modularity in deep learning solutions can bring a number of advantages to deep learning solution development that other more  traditional engineering domains already experience. Fig. \ref{fig:modularity_banner_image} shows a visual representation of monolithic ML solution development in comparison to modular ML solution development.
% As a first step towards this direction we are exploring the cost benefit trade offs of solving machine learning solutions in a more modular way.
As a first step towards this end, we explore the cost benefit trade-offs of solving machine learning problems in a multi stage modular way in comparison to solving them with a monolithic deep learning solution.
% We do it through several case studies. 
In this work, we perform this analysis through three example problems. For each example problem, we will implement deep learning solutions in two ways. One of the solutions will be developed taking a monolithic approach. The other solution will be developed taking a modular approach. Then, we will be evaluating the two approaches quantitatively and qualitatively to analyze the trade-offs of each approach. One of the objectives of our experiments is to study if ML modules can demonstrate modularity characteristics similar to traditional non-ML software modules.
% What costs and benefits will be studied? Dataset acquisition and availability, model development {Model development effort:[Reusability, Learning rate], Accuracy and efficiency, maintainability}. Are there measures to quantify these?
During the evaluation we will focus on several aspects of monolithic and modular solution development. 1/ Exploring how the two approaches will be affected by dataset availability and cost of data acquisition. 2/ Study the differences of solution development effort between monolithic approach and the modular approach. 3/ Trade offs between model performance in terms of accuracy and efficiency. 4/ Maintainability of the solutions developed with the two approaches.

%%%%%%%%%%%%%%%%%%%%%%%%%%%%%%%%%%%%%%%%%%%%%%%%%%%%%%%
%%%%%%%%%%%%%%%%%%%%%%%%%%%%%%%%%%%%%%%%%%%%%%%%%%%%%%%
%%%%%%%%%%%%%%%%%%%%%%%%%%%%%%%%%%%%%%%%%%%%%%%%%%%%%%%
%%%%%%%%%%%%%%%%%%%%%%%%%%%%%%%%%%%%%%%%%%%%%%%%%%%%%%%
%%%%%%%%%%%%%%%%%%%%%%%%%%%%%%%%%%%%%%%%%%%%%%%%%%%%%%%
%%%%%%%%%%%%%%%%%%%%%%%%%%%%%%%%%%%%%%%%%%%%%%%%%%%%%%%
%%%%%%%%%%%%%%%%%%%%%%%%%%%%%%%%%%%%%%%%%%%%%%%%%%%%%%%

\section{Modularity in Machine Learning}
% Define modularity. What are the expectations of modules? Expected benefits.

% An introduction to Modularity in ML. What and why?
Modularity is a familiar approach when solving complex problems in many domains. It simplifies the problem solving process by breaking down a complex problem into more manageable subparts. After breaking the problem into subproblems the subproblems can be solved independently. This process brings a number of advantages to the solution engineering process. The goal of making ML solutions modular is to enable these advantages to ML solution engineering.

% Solution development benefits.
%     Reusability {data and models}
\textbf{Combinatorial generalization} is one of the main advantages of modular ML models. In general an ML model that is trained for a specific problem is only useful for that specific problem. Therefore the monolithic models that can not be broken down into submodules are not usable outside of the problem that it intends to solve. On the other hand modular ML models are composed of more than one ML model that each model solves a sub problem of the original problem. This allows parts of the modular ML models to be reused in different contexts beyond the original problem. This enables mixing and matching modules to create new unique models to solve different problems. This helps to minimize the chances of having to start from the scratch when developing unique solutions. On the other hand, when using end to end learning methods that learn a monolithic model, very often, each new problem is a unique problem that has to be solved starting from fundamental technologies.

The same is true for datasets that are used to train ML models. Training monolithic models require a dataset specific for the intended problem that each datapoint maps a specific input type to a specific target type with other required characteristics. Such a dataset is only useful to train a model similar to the original problem. In comparison, modular ML models are a composition of submodules where each model is trained with a dataset that matches the subproblem. At the same time, the process of breaking down a larger problem into subproblems usually results in subproblems that are simpler and more general. Therefore, within a development ecosystem, modular ML solutions increase the chances of reusing datasets to solve different problems.

%     Domainexpert intervention.
\textbf{Domain expert intervention} to simplify the learning process by incorporating domain knowledge when developing solutions. One of the ways to do this is by decomposing the problem in a way that the resulting subproblems are more data efficient and simpler to learn. As a simple example, think about a multi-digit classification problem. In this case, a domain expert may break down the problem to detect each digit individually using a simpler model and later aggregate the classification results to find the classification for the original multi digit input. This simple decomposition made the subproblem simpler as well as more data efficient while increasing the amount of data points per classification class.
%     Parallel development like in SE.
In traditional software engineering modularity allows engineering teams to divide work across individuals or specialized teams. This minimizes the development overhead by assigning units of work with cross-cutting concerns to one individual or one team. In this process a module boundary can define a unit of work that can be assigned to an individual or a team. In addition to that this helps to reduce dependencies among units of work and parallelize the development process.

% Performance characteristics.
\textbf{Performance tuning ability} is higher with modular ML models. Modularity enables breaking down a problem into subproblems and addressing them separately. Since subproblems tend to be more general than the overall high level problem, finding technologies and existing solutions for the subproblems is likely to be easier. This enables more options for solution developers to make performance trade-offs at the subproblem level. Further, modularity makes it easier to do incremental improvements to solutions. Individual modules in modular systems can be replaced with different modules with the same functionality but with different characteristics. As an example, one may trade off accuracy for faster performance at one subproblem of the modular solution to meet a business requirement at hand. With monolithic solutions, making performance trade-offs usually require replacing the whole model with a one that has the required characteristics.

% Maintainability.
\textbf{Maintainability} of modular ML models is higher compared to monolithic ML models. Since the submodules of modular ML models can be replaced with different models with the same functionality, newer or improved technologies emerge, submodules can be upgraded without requiring major changes to the entire solution. In addition to that, since modular models are a composition of semantically meaningful models, they are more human understandable. This opens up more opportunities to verify and monitor modular models. This simplifies the process of isolating issues and troubleshooting.

% Economies of scale.
\textbf{Economies of scale} effect can be harnessed better with modular ML models from a development ecosystem perspective. Since modular ML models can take better advantage of existing ML modules by reusing them to create different higher level solutions, the modules that are more commonly reused in many problems get a higher demand from the development community. As a result of this, more demanding modules are likely to be further improved within the ecosystem due to the economies of scale effect and these improvements can be exploited by the downstream solutions. On the other hand, in end to end ML, this effect is relatively less prominent since the learned solutions are monolithic and specialized to individual problems making them less usable in other contexts.

% Move to case studies.
% Why does ML not care about this?
% Compare them with monolithic versions.
% Introduce a set of modularity aspects that we are interested in. Then in related work show that other works do not satisfy these aspects.

%%%%%%%%%%%%%%%%%%%%%%%%%%%%%%%%%%%%%%%%%%%%%%%%%%%%%%%
%%%%%%%%%%%%%%%%%%%%%%%%%%%%%%%%%%%%%%%%%%%%%%%%%%%%%%%
%%%%%%%%%%%%%%%%%%%%%%%%%%%%%%%%%%%%%%%%%%%%%%%%%%%%%%%
%%%%%%%%%%%%%%%%%%%%%%%%%%%%%%%%%%%%%%%%%%%%%%%%%%%%%%%
%%%%%%%%%%%%%%%%%%%%%%%%%%%%%%%%%%%%%%%%%%%%%%%%%%%%%%%
%%%%%%%%%%%%%%%%%%%%%%%%%%%%%%%%%%%%%%%%%%%%%%%%%%%%%%%
%%%%%%%%%%%%%%%%%%%%%%%%%%%%%%%%%%%%%%%%%%%%%%%%%%%%%%%
% Introduce the example problems.

\section{Case Studies}

As discussed before, we will be using three example problems in order to empirically highlight the benefits and trade-offs of modular and monolithic machine learning solutions. First example problem is a text based sentiment analysis. The other two problems are a satellite image classification problem and a near infrared (NIR) field prediction problem. 

\subsection{Text based Sentiment Analysis}

% We use sentiment analysis given a language.
Text based sentiment analysis is useful in a number of business contexts. For instance, businesses use sentiment analysis to understand consumer sentiment towards their brand and the products. In today's internet based global market setting, analyzing the sentiment of a text in a given language is important for many business organizations. In this section we are using this problem as a proxy to study the trade-offs of monolithic and modular machine learning solutions.

% Explain the monolithic and modular approaches for this problem
Deep learning technologies have demonstrated state of the art performance in sentiment analysis tasks. Current prevalent deep learning based models are monolithic in nature \cite{DBLP:conf/icml/TanL19, DBLP:conf/naacl/DevlinCLT19}.
Monolithic deep learning solutions for sentiment analysis train machine learning models end to end to predict the sentiment of a text from a given source language. More modular solution for sentiment analysis is to develop the solution using two modules that solve the problem in two intuitive stages. The first stage is to translate the source language text to a suitable target language. The second stage is to analyze the sentiment of the translated text. This approach enables the opportunity to train a sentiment analysis model for a language with a larger and more representative sentiment analysis dataset or to find a sentiment analysis model already trained for a specific language that demonstrates higher performance characteristics. In this section we will be implementing two sentiment analysis solutions one monolithic and one modular as described before and compare the advantages and disadvantages of the two approaches.

\begin{figure}[ht]
    \centering
    \includegraphics[width=0.46\textwidth]{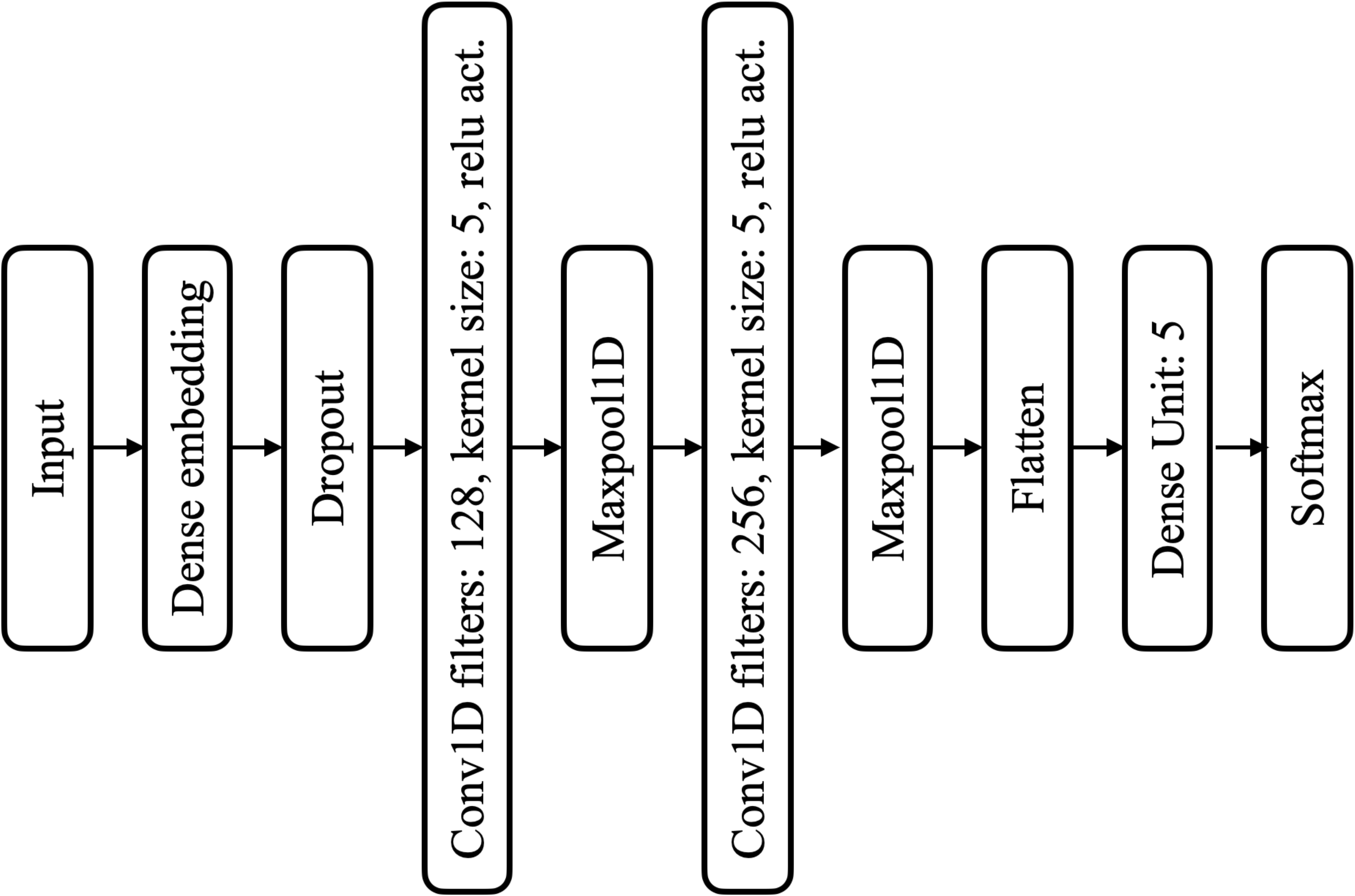}
    \caption{Architecture of the sentiment analysis student model.}
    \label{fig:archi_sentiment_distil}
\end{figure}

% Explain the experiment and its setup. This will include diagrams explaining the setup.
This experiment is performed with a Spanish sentiment analysis problem. Off the shelf pretrained language models are used to implement the monolithic solution and the two stage modular solution for this problem.
% nlptown/bert-base-multilingual-uncased-sentiment
The monolithic version of the experiment is conducted using an existing BERT based pretrained multilingual sentiment analysis model \cite{Wolf2019HuggingFacesTS}. The model is already trained with 150k English, 80k Dutch, 137k German, 140k French, 72k Italian and 50k Spanish sentiment analysis data points. The model predicts the sentiment of the input text in a 1 to 5 scale where 1 is the least positive and 5 is the most positive sentiment.
% Helsinki-NLP/opus-mt-es-en
The first stage of the modular version is implemented with an existing pretrained spanish to english translation model \cite{Wolf2019HuggingFacesTS, tiedemann-2020-tatoeba}. The second stage uses the same sentiment analysis model that is used in the monolithic version. 
% Distilled versions (of both mono and modu).
Further, the monolithic solution and the modular solution were distilled~\cite{DBLP:journals/corr/HintonVD15} into a smaller model to see the performance characteristics of the distilled solutions in comparison.
% Distillation-compilation analogy.
At a high level, the distillation process is analogous to program compilation in software development. In software development, the result of compilation is an object that runs much faster during deployment. However, the compiled object is less meaningful to humans compared to the program written using a high level programming language. Similarly, in the case of a modular ML solution, the distillation results in a faster solution but compromises the explainability that is present in the modular solution.
% Distillation process.
Distillation is performed with a smaller convolutional architecture as the student network in both monolithic and modular cases. Input to this model is an integer sequence that was created using a word dictionary that contains a unique index for each word in the corpus. Inputs to the student models are truncated to a maximum length of 500 words and padded appropriately if a sequence is short. The architecture of the distillation student network is shown in fig. \ref{fig:archi_sentiment_distil}.
% Add more information about the distillation model and the loss.
The distillation is done by considering the teacher model as a blackbox to keep the process more generally applicable. In the distillation process the student is trained to imitate the teacher by minimizing the cross entropy loss between the teacher model's output and the student model's output. This distillation process does not require any labeled data points. It only needs unlabeled text from the input language.

\begin{figure}[ht]
    \centering
    \includegraphics[align=c,width=0.22\textwidth]{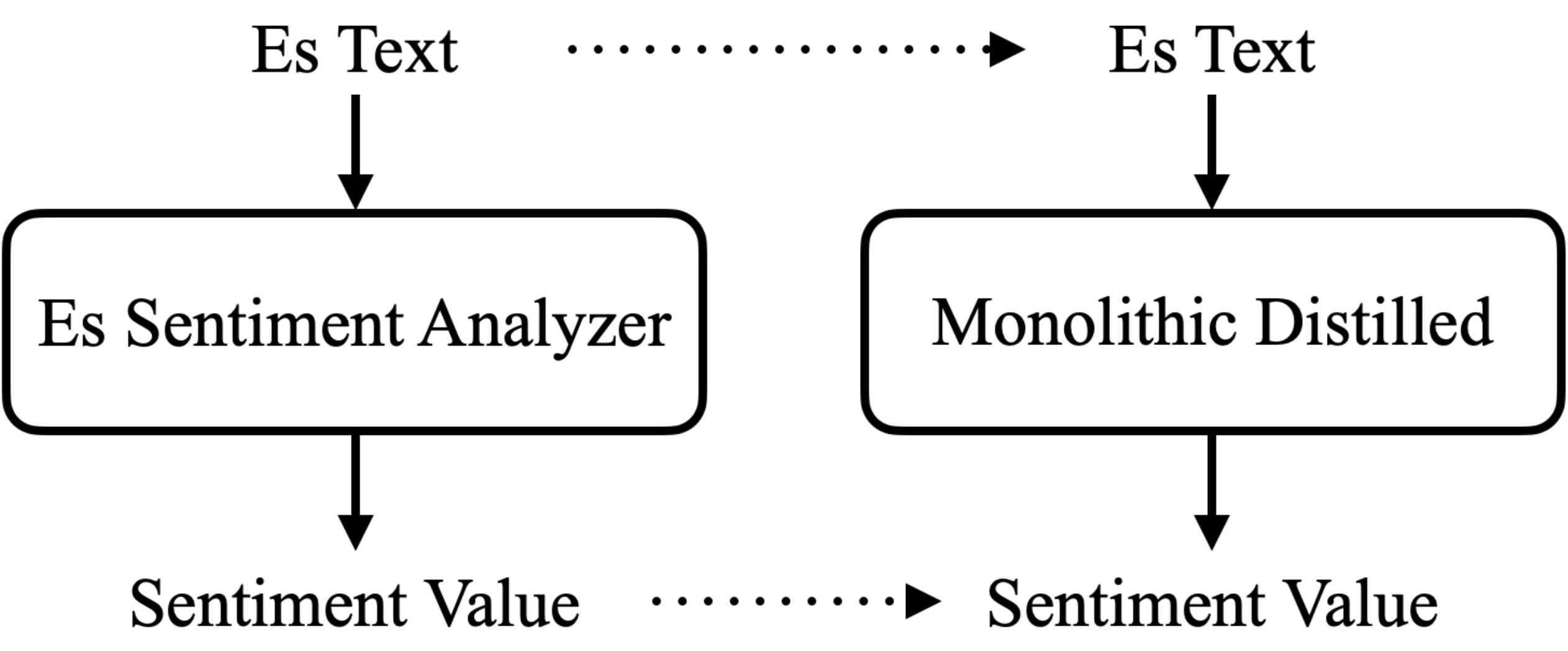} \hspace{8pt}
    \includegraphics[align=c,width=0.22\textwidth]{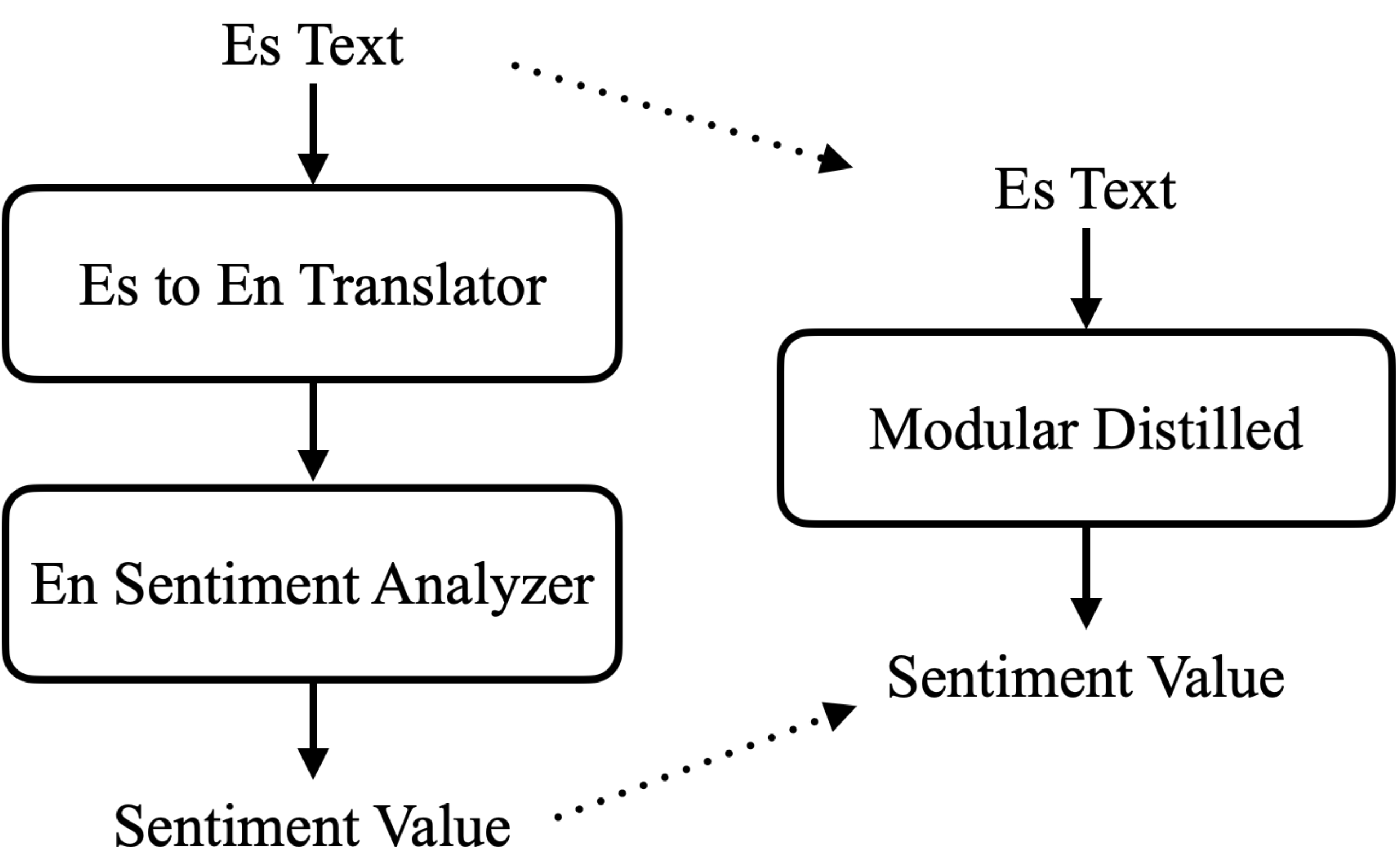}
    \caption{Monolithic (left) vs modular (right) sentiment analysis models and the models distilled from them.}
    \label{fig:mono_vs_modu_sentim_with_distilled}
\end{figure}

Fig. \ref{fig:mono_vs_modu_sentim_with_distilled} shows a diagram of the monolithic and the modular solutions used in the experiment with their distilled counterparts.

% Dataset.
Performance characteristics of these models are compared using the test set of the spanish portion of the amazon multilingual product review sentiment dataset \cite{DBLP:conf/emnlp/KeungLSS20}. This contains 30000 sentiment analysis data points. Each datapoint has spanish product review text and a 1 to 5 star rating that correspond to the product review text.

\subsection{Satellite Image Classification and NIR prediction}

% Content structure will be similar to the previous experiments. 
% However, this is a different ML method and a different medium as well (language vs imaging).
% This experiment is supposed to solidify the observations of the previous experiments and hint the generality.
% This is an experiment done with custom trained models.
% Content structure will be similar to the previous experiments. 
% However, this is a different ML method and a different medium as well (language vs imaging).
% This experiment is supposed to solidify the observations of the previous experiments and hint the generality.

% Todo:
% Change the denoise split to 70%.
% Motivation:
% 	The picture is bigger than just getting a model design advantage due to modularity. This is an opportunity to gather forces. Make an ecosystem. Focus the work of a lot of people on something narrow to get the best out of it. Work not as one loney team but as the whole of humanity on projects.
% You can mathematically write about cross entropy minimization (imitating one distribution from another) used for the distillation.

% This is an experiment done with satellite images.
Satellite image based remote sensing is useful in a number of real world applications such as land survey, surveillance, traffic monitoring etc. In this section we are studying the trade-offs between modular vs monolithic ML solutions using a satellite image classification problem. In this problem we are classifying cloudy satellite images based on the EuroSAT dataset \cite{DBLP:journals/staeors/HelberBDB19} into 10 classes of land use and land cover. We will be implementing one monolithic model and two modular models for this classification problem. Performance of these three models will then be evaluated based on accuracy and latency. The accuracy of the solutions are also compared with a large percentage of weight pruning. Further, we evaluate the performance of models distilled from the monolithic and two modular models. Next, we reuse a module from the modular classification solutions to predict the near infrared (NIR) band of the EuroSAT based cloudy satellite images. This modular NIR band prediction model is compared with a monolithic model for the same task.
% Custom trained models.
Unlike in the previous example, in this example we are training custom ML models for the problem.

\begin{figure}[ht]
    \centering
    \includegraphics[width=0.48\textwidth]{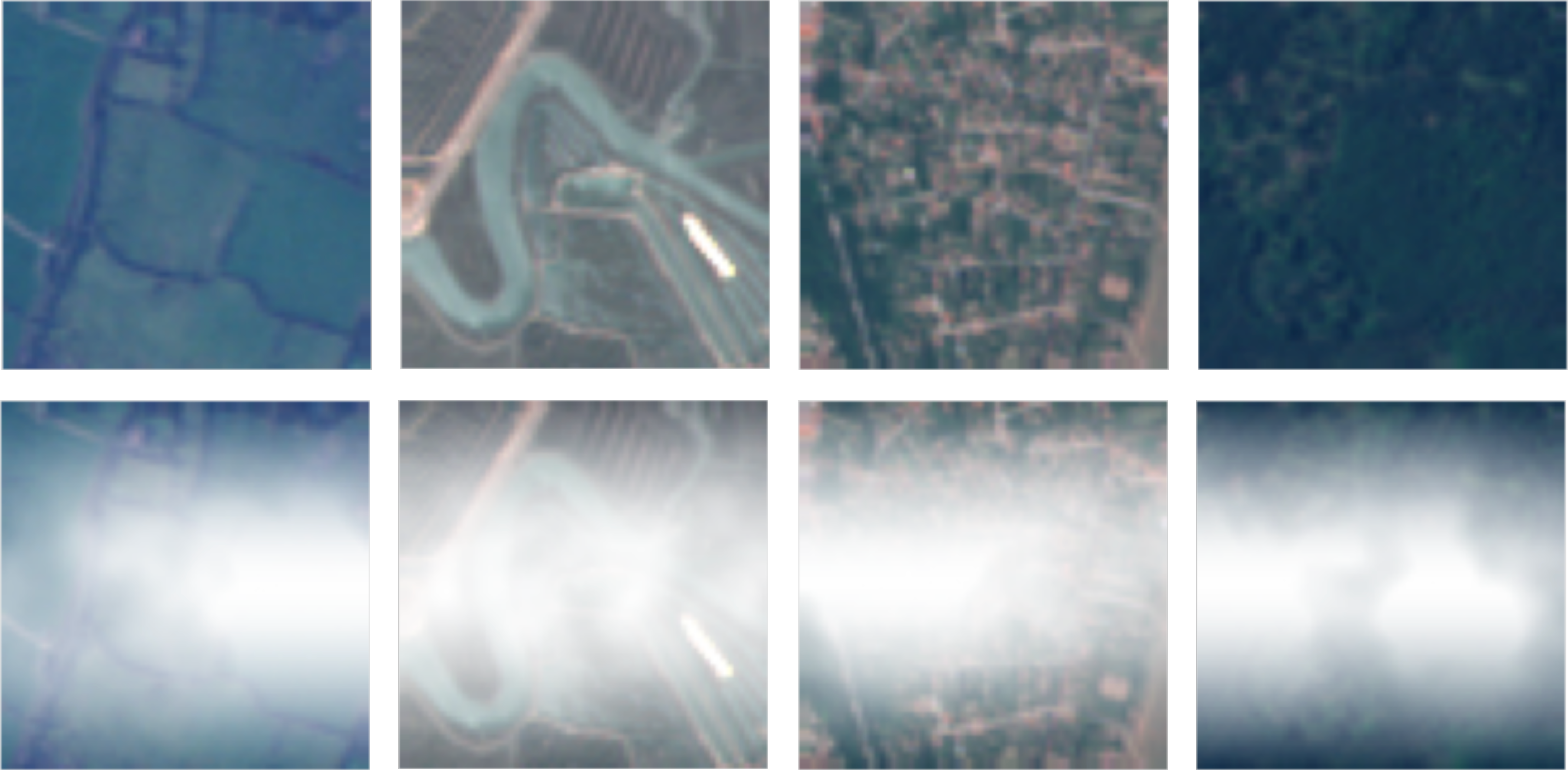}
    \caption{EuroSAT data points before (top row) and after (bottom row) adding cloud layers.}
    \label{fig:data_eurosat_vs_cloud_added}
\end{figure}

% Dataset details. Bands. Splits. Adding clouds.
EuroSAT dataset has a red green blue (RGB) version and a 13 band version. For our experiments, we use the RGB and NIR bands in the EuroSAT dataset. EuroSAT dataset only contains clear satellite images without obstructions. We alter this dataset by adding a cloud overlay to the RGB bands using the approach proposed by Kenji et al. \cite{DBLP:conf/cvpr/EnomotoSWFMNK17}. Fig. \ref{fig:data_eurosat_vs_cloud_added} shows a sample of the altered EuroSAT dataset.
% Image after adding clouds to EuroSAT
Added clouds makes the dataset more challenging for prediction tasks compared to the original EuroSAT dataset. Since we want to represent the low labeled data regime that is commonly seen in real world industrial application domains, we use only a 20\% of the EuroSAT labeled data to train, validate and test the models for the classification and NIR band prediction tasks. In the case of the classification task we use the cloudy RGB image as the input and the corresponding classification label as the target. In the case of NIR prediction, the cloudy RGB image is used as the input and the corresponding NIR band is used as the target. The rest of the data that is not used for the classification and NIR prediction is used to train, validate and test the cloud removal module that is used in the modular models. It should be noted that this training step does not utilize the labels from the original EuroSAT dataset. This cloud removal dataset has the RGB images with the cloud overlay as the input and the corresponding cloud free RGB images as the target. Such an unlabeled dataset is relatively easy to acquire in larger quantities in practice in the real world as well since it does not involve manual data labeling.

\begin{figure}[ht]
    \centering
    \includegraphics[width=0.48\textwidth]{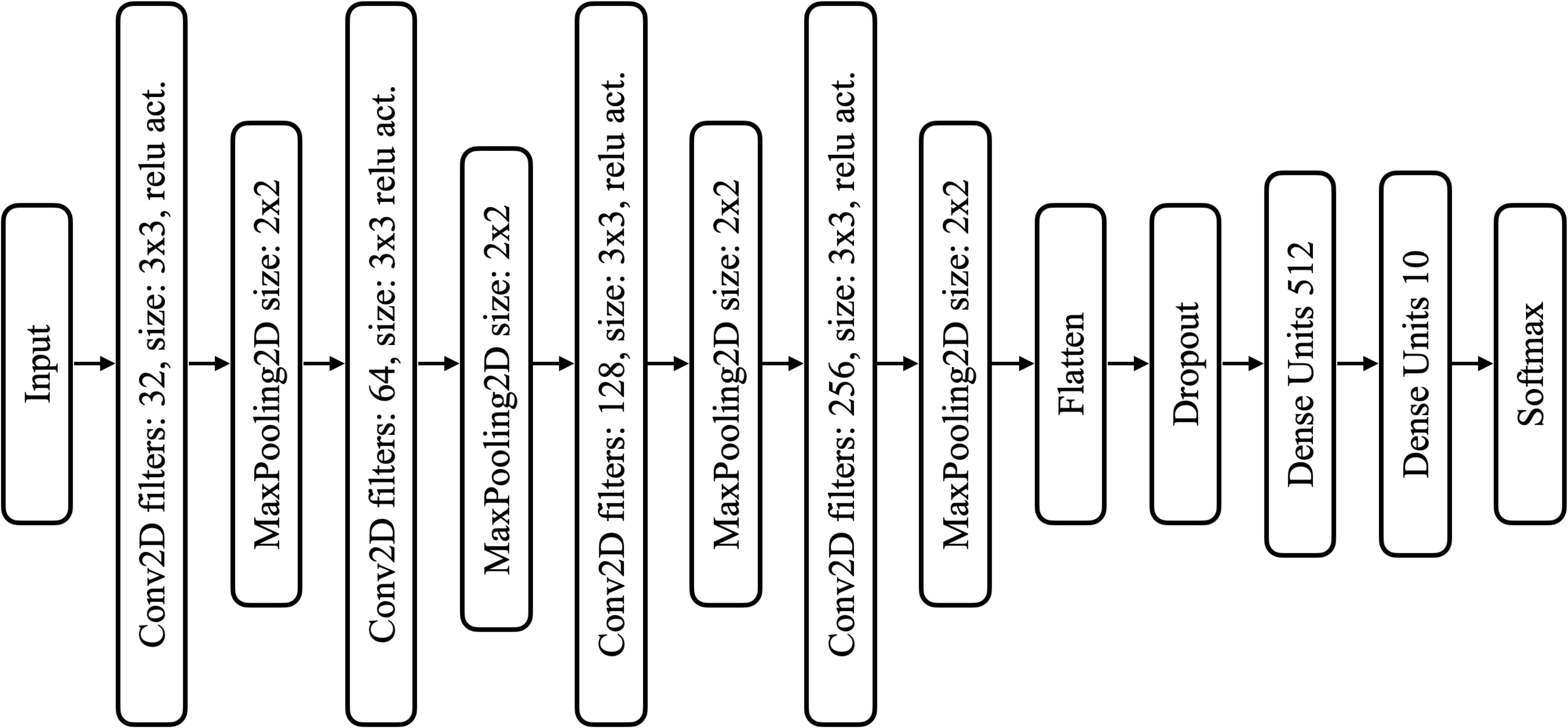}
    \caption{Architecture of the satellite image classification model.}
    \label{fig:diag_archi_eurosat_classif}
\end{figure}

\begin{figure}[ht]
    \centering
    \includegraphics[width=0.48\textwidth]{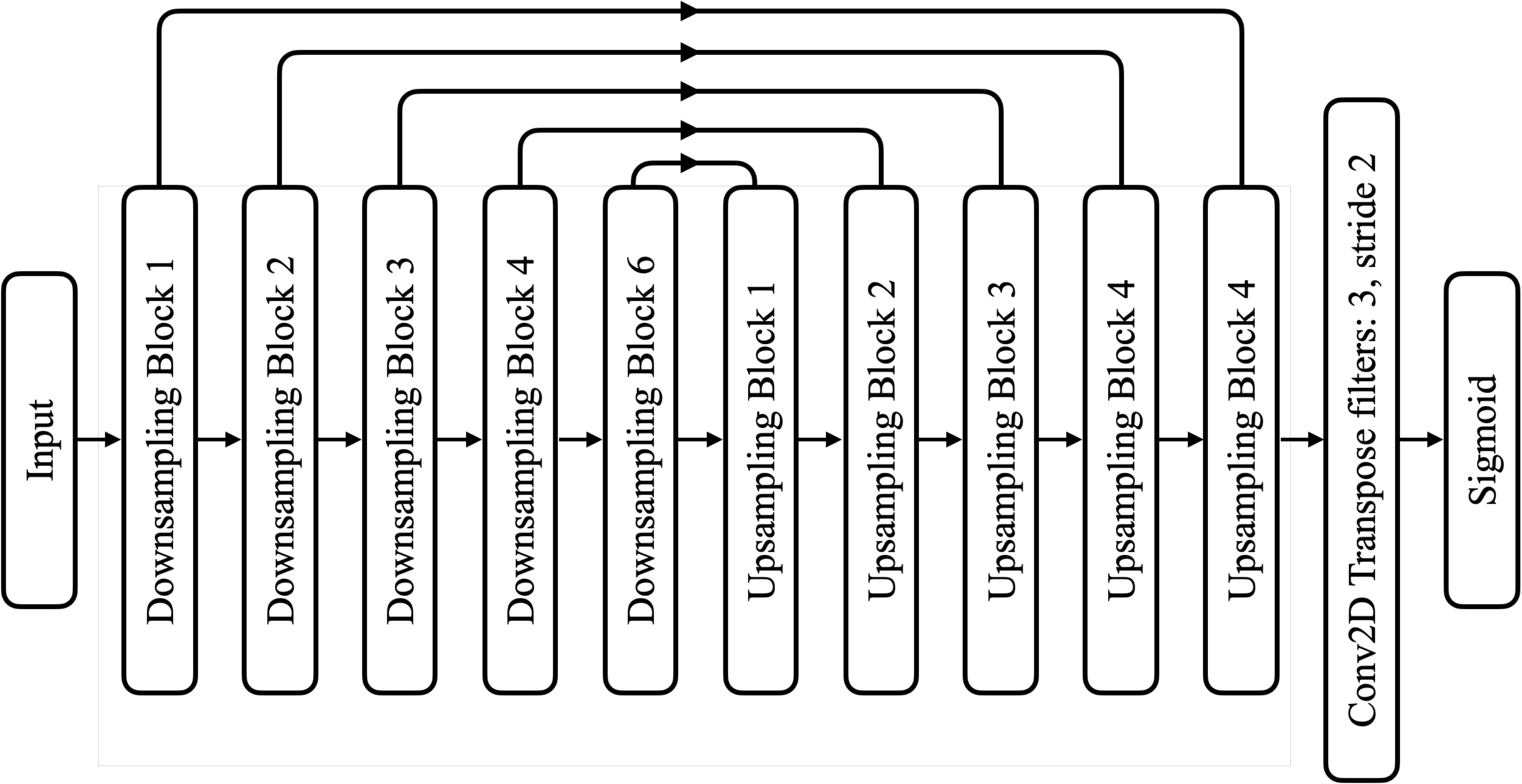}
    \caption{Encoder decoder architecture of the cloud removal model.}
    \label{fig:diag_archi_eurosat_cloud_remover}
\end{figure}

\begin{figure}[ht]
    \centering
    \includegraphics[width=0.48\textwidth]{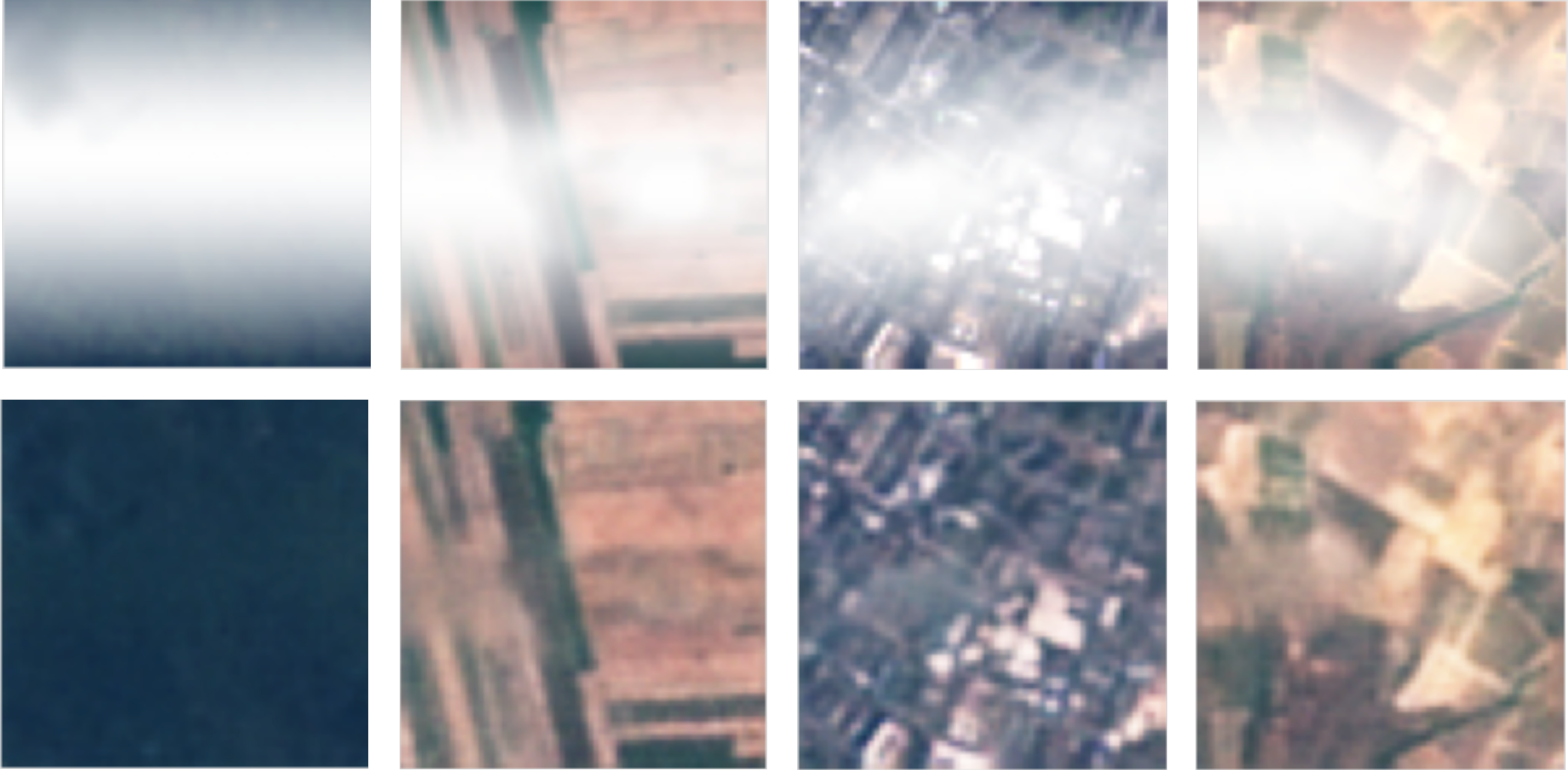}
    \caption{Output (bottom row) of the cloud removal module when it is fed with cloudy data (top row).}
    \label{fig:data_cloudy_vs_cloud_removed}
\end{figure}

% Describing the three solutions that we are going to compare.
Our monolithic classification model is trained end to end, validated and tested using the classification split of the cloudy satellite images. The network architecture used in the monolithic model is shown in fig. \ref{fig:diag_archi_eurosat_classif}.
% Classifier architecture (image) explanation.
This architecture downsamples the feature maps while increasing the number of channels as the layers progress from input to the output. In the final layers the output of the convolutional layers are flattened and sent through dense layers to do the classification. Dropouts are used before the features are fed to the dense units.
% Fig of monolithic architecture.
The modular versions perform the classification in two stages. The first stage performs cloud removal. Cloud removal is performed using the encoder decoder network architecture that is shown in fig. \ref{fig:diag_archi_eurosat_cloud_remover}.
% Cloud remover architecture (image) explanation.
First part of this architecture, the encoder part, downsamples the feature maps using 6 downsampling blocks. Each downsampling block has a convolutional layer, batch normalization layer and a leaky relu activation. The first downsampling block does not have batch normalization. Each downsampling block from input to output reduces the feature map size while increasing the number of channels. The second part of the architecture, the decoder part, upsamples the feature maps that are downsampled by the decoder using four upsampling blocks. Each upsampling block has a convolutional transpose layer, dropout layer and a relu activation layer. The first upsampling block does not have a dropout layer. Each upsampling layer from input to output increases the width and the height of feature maps while reducing the number of channels. As shown in the architecture figure, skip connections are used from the downsampling blocks to upsampling blocks with a mirror-like correspondence to incorporate low level features to the latter upsampling layers. Finally the output of the upsampling layers are sent through another convolutional transpose layer with three channels and a sigmoid activation function. This architecture is adapted from the U-net \cite{DBLP:conf/miccai/RonnebergerFB15} and pix2pix \cite{DBLP:conf/cvpr/IsolaZZE17} architectures.
% Fig of cloud remover architecture.
The second stage performs classification on the cloud removed images. In the first modular solution the classifier is trained using the output of the cloud removal module. This modular version represents the case where the classification data available for training is cloudy. We will be calling this solution modular (I). In the second modular version the classifier is trained using cloud free satellite images. This modular version represents the case where the classification data available for training are cloud free. We will be calling this solution modular (II). This corresponds to making a barebone pretrained satellite image classification module to be published in a model repository to be used by others. All classifiers have the same network architecture shown in fig. \ref{fig:diag_archi_eurosat_classif} and they are trained using the classification data split that was discussed before. The cloud removal module is trained with the remaining data split that was described before. The output of the cloud removal module is shown in fig. \ref{fig:data_cloudy_vs_cloud_removed}.
% Output of cloud removal module.
% Accuracy experiment.
Further, three distilled models are trained from each monolithic, modular (I) and modular (II) solutions. Distillation is done by using the same approach that was used in the sentiment analysis case. However, before the distillation, student models that correspond to the monolithic solution and the modular (I) version are pre-trained with the available labeled cloudy classification data.
% Todo: Maybe pretrain modular (II) distillation student model with clean classification data.
Student model of the modular (II) version is not pretrained this way because it represents the case where cloudy labeled classification data is not available.

%%%%%%%%%%%%%%%%%%%%%%%%%%%%%%%%%%%%%%%%%%%%%%%%%%%%%%%
%%%%%%%%%%%%%%%%%%%%%%%%%%%%%%%%%%%%%%%%%%%%%%%%%%%%%%%
%%%%%%%%%%%%%%%%%%%%%%%%%%%%%%%%%%%%%%%%%%%%%%%%%%%%%%%
%%%%%%%%%%%%%%%%%%%%%%%%%%%%%%%%%%%%%%%%%%%%%%%%%%%%%%%
%%%%%%%%%%%%%%%%%%%%%%%%%%%%%%%%%%%%%%%%%%%%%%%%%%%%%%%
%%%%%%%%%%%%%%%%%%%%%%%%%%%%%%%%%%%%%%%%%%%%%%%%%%%%%%%
%%%%%%%%%%%%%%%%%%%%%%%%%%%%%%%%%%%%%%%%%%%%%%%%%%%%%%%

\section{Trade-offs of Modularity}

In this section we will use the solutions we developed for the three example problems to compare and contrast the trade-offs between monolithic and modular solutions.

%%%%%%%%%%%%%%%%%%%%%%%%%%%%%%%%%%%%%%%%%%%%%%%%%%%%%%%
%%%%%%%%%%%%%%%%%%%%%%%%%%%%%%%%%%%%%%%%%%%%%%%%%%%%%%%
%%%%%%%%%%%%%%%%%%%%%%%%%%%%%%%%%%%%%%%%%%%%%%%%%%%%%%%
%%%%%%%%%%%%%%%%%%%%%%%%%%%%%%%%%%%%%%%%%%%%%%%%%%%%%%%
%%%%%%%%%%%%%%%%%%%%%%%%%%%%%%%%%%%%%%%%%%%%%%%%%%%%%%%
%%%%%%%%%%%%%%%%%%%%%%%%%%%%%%%%%%%%%%%%%%%%%%%%%%%%%%%
%%%%%%%%%%%%%%%%%%%%%%%%%%%%%%%%%%%%%%%%%%%%%%%%%%%%%%%

\subsection{Accuracy}

This section compares the accuracy of the monolithic and modular solutions using solutions developed for the sentiment analysis and satellite image classification problems.

%%%%%% Sentiment.
\subsubsection{Sentiment Analysis}

\begin{figure}[ht]
    \centering
    \includegraphics[width=0.46\textwidth]{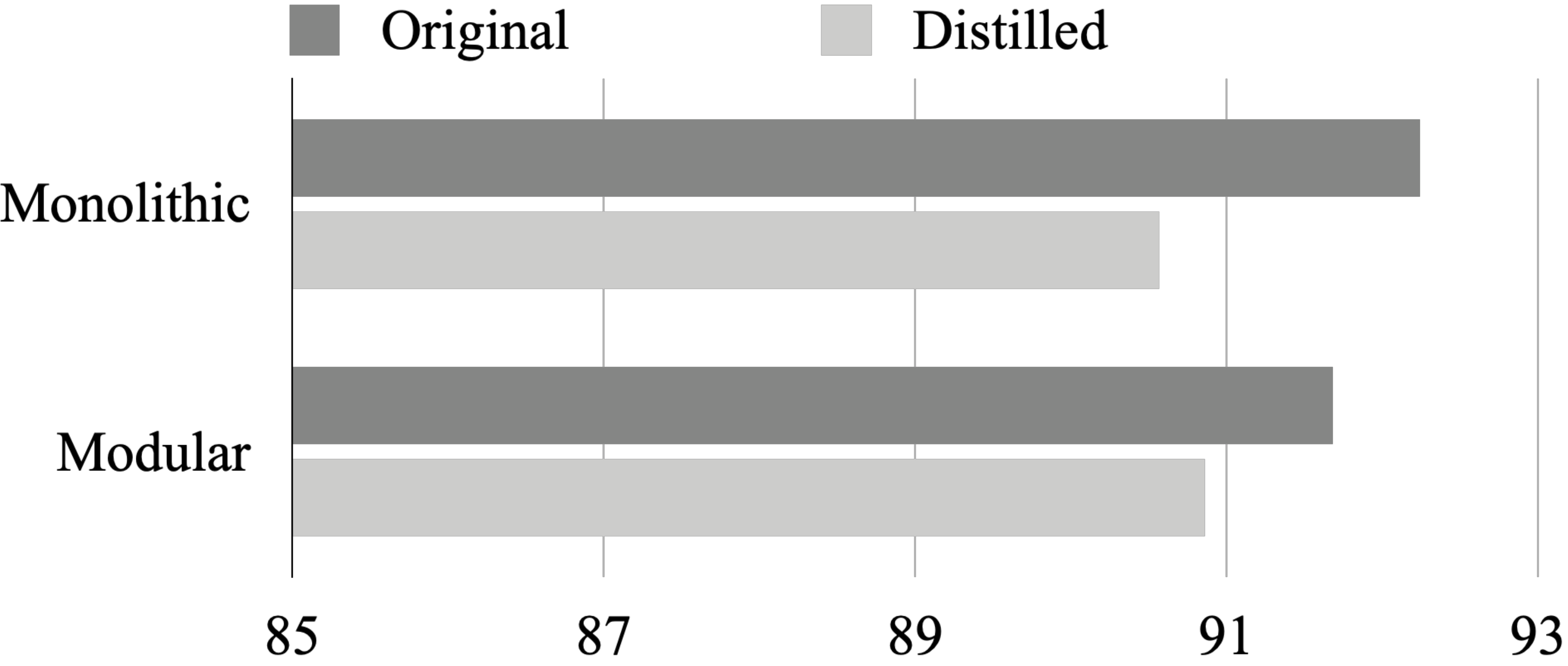}
    \caption{Sentiment analysis accuracy of monolithic and modular solutions including their distilled versions.}
    \label{fig:accuracy_of_mono_vs_modu_with_distilled_senti}
\end{figure}

% Accuracy experiment.
For the sentiment analysis case, the spanish product review test set from the amazon multilingual product review sentiment dataset is used to compare the accuracy of the two original monolithic and modular solutions and the two distilled versions of the solution. In this experiment we use a one-off accuracy measure to quantify the performance. Here we consider a prediction of the model as correct if the predicted star rating is exact or off by only one. If not, we consider the prediction as incorrect. We believe that this measure is more realistic because there is no universally agreeable star rating for a given review text. The results of this experiment are shown in fig. \ref{fig:accuracy_of_mono_vs_modu_with_distilled_senti}.
% Provide an explanation to the results.

The results in fig. \ref{fig:accuracy_of_mono_vs_modu_with_distilled_senti} shows that the original monolithic solution has less than 1 percentage point higher accuracy over the original modular solution. The solution distilled from the monolithic solution is less than 2 percentage points lower in accuracy compared to the original monolithic solution. The model distilled from the modular solution is less than 1 percentage point lower compared to the original modular solution. The model distilled from the modular version has 0.3\% higher accuracy compared to the model distilled from the monolithic solution. The results show that modular solutions can be comparable in terms of accuracy to monolithic solutions.

%%%%%% Satellite.
\subsubsection{Satellite Image Classification}

\begin{figure}[ht]
    \centering
    \includegraphics[width=0.46\textwidth]{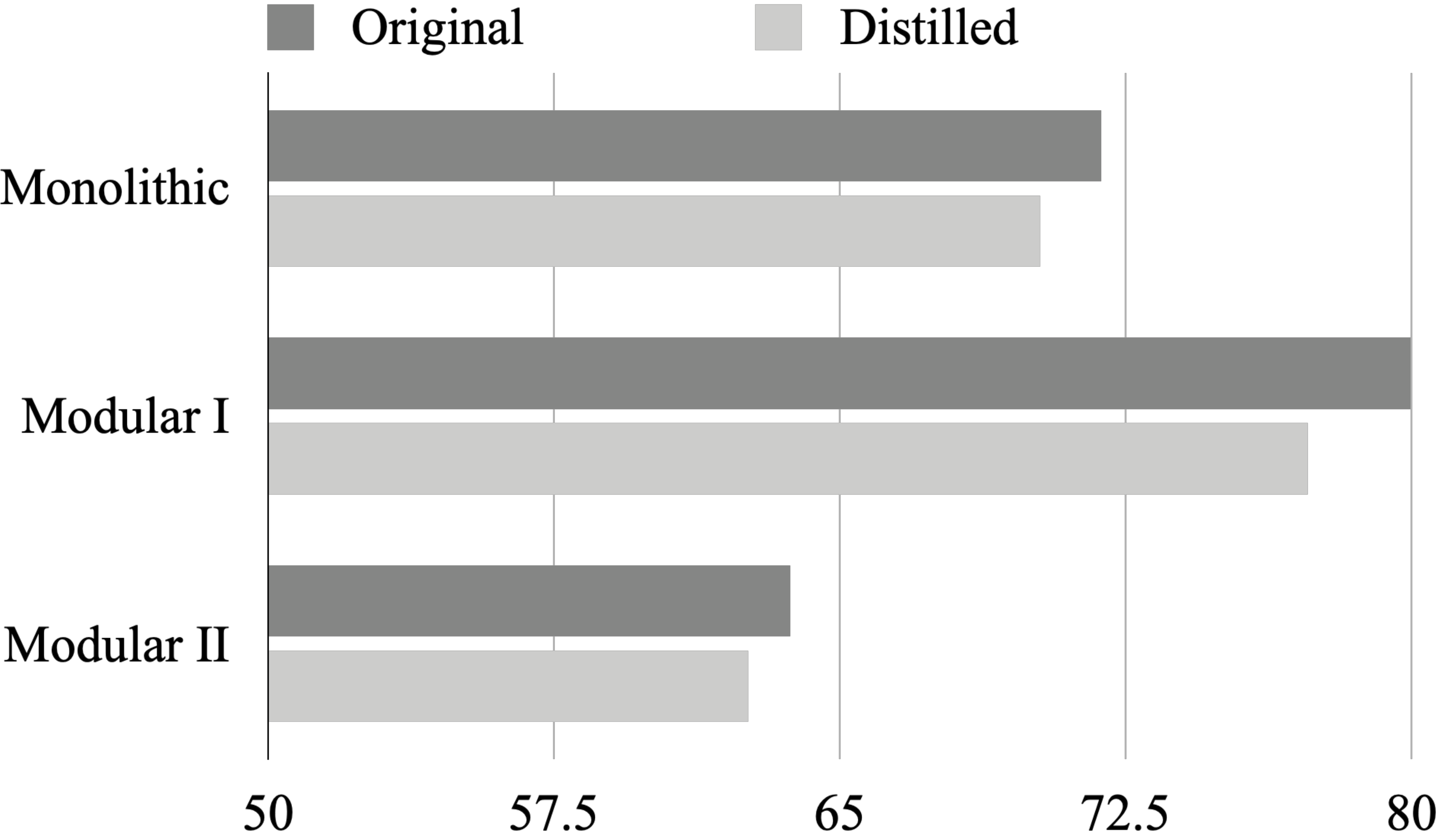}
    \caption{Cloudy satellite image classification accuracy of monolithic and modular solutions including their distilled versions.}
    \label{fig:chart_mono_modu.i_modu.ii_distilled_sat_accuracy}
\end{figure}

The test split of the EuroSAT dataset is used to measure the accuracy of the models and the results are shown in fig. \ref{fig:chart_mono_modu.i_modu.ii_distilled_sat_accuracy}.

The results in fig. \ref{fig:chart_mono_modu.i_modu.ii_distilled_sat_accuracy} shows that the modular (I) solution that was trained with cloudy labeled data with the added cloud removal module performs the best out of the three solutions. It has been able to achieve 11.34\% accuracy improvement over the monolithic solution when comparing the original non-distilled solutions. When comparing the distilled solutions, the modular (I) solution was able to achieve 10\% accuracy improvement over the corresponding monolithic solution. This accuracy improvement can be attributed to the cloud removal module in the modular version that was capable of utilizing low cost unlabeled data. Modular (II) solution in which the classification module is trained with clean data has the lowest accuracy. However, this solution attains its accuracy level without using any labeled data points from its target problem, cloudy satellite image classification.

%%%%%%%%%%%%%%%%%%%%%%%%%%%%%%%%%%%%%%%%%%%%%%%%%%%%%%%
%%%%%%%%%%%%%%%%%%%%%%%%%%%%%%%%%%%%%%%%%%%%%%%%%%%%%%%
%%%%%%%%%%%%%%%%%%%%%%%%%%%%%%%%%%%%%%%%%%%%%%%%%%%%%%%
%%%%%%%%%%%%%%%%%%%%%%%%%%%%%%%%%%%%%%%%%%%%%%%%%%%%%%%
%%%%%%%%%%%%%%%%%%%%%%%%%%%%%%%%%%%%%%%%%%%%%%%%%%%%%%%
%%%%%%%%%%%%%%%%%%%%%%%%%%%%%%%%%%%%%%%%%%%%%%%%%%%%%%%
%%%%%%%%%%%%%%%%%%%%%%%%%%%%%%%%%%%%%%%%%%%%%%%%%%%%%%%

\subsection{Latency}

This section compares the latency of the monolithic and modular solutions using solutions developed for the sentiment analysis and satellite image classification problems.

%%%%% Sentiment.
\subsubsection{Sentiment Analysis}

\begin{figure}[ht]
    \centering
    \includegraphics[width=0.46\textwidth]{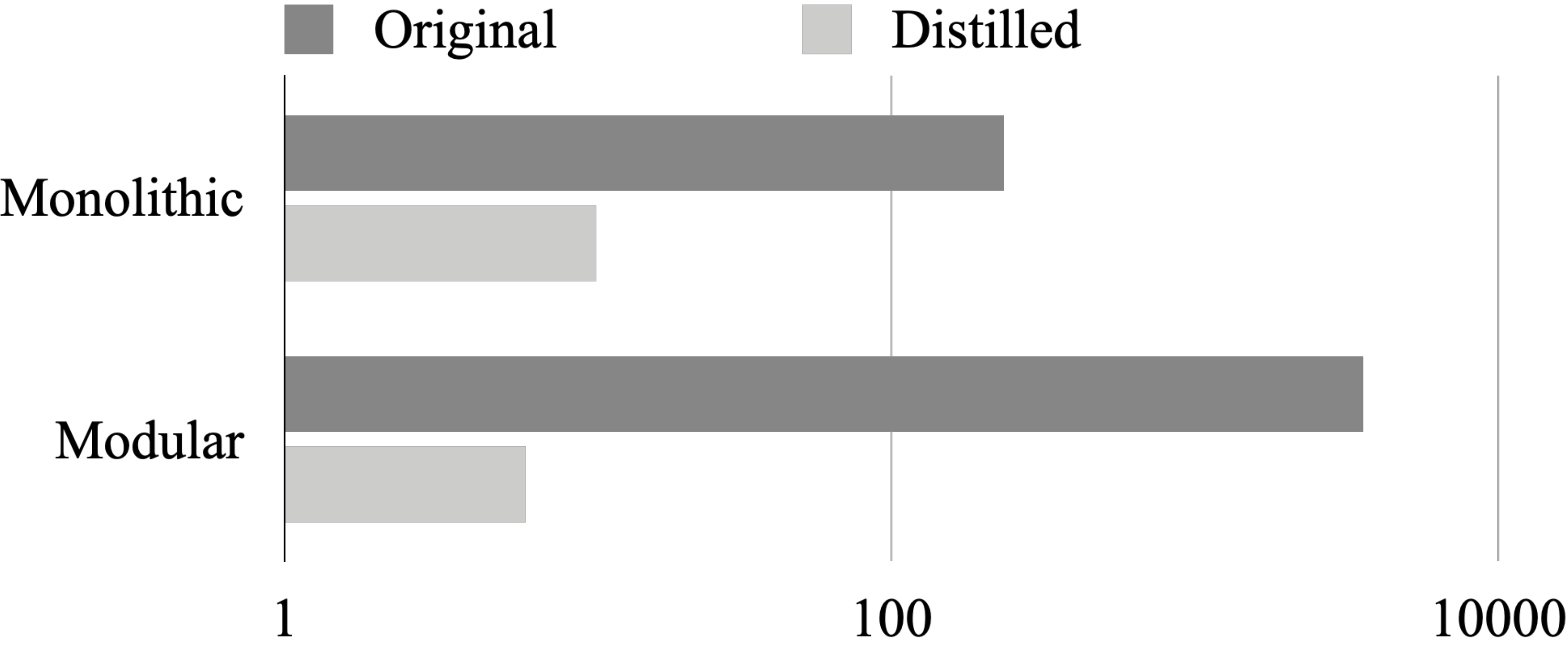}
    \caption{Sentiment analysis latency of monolithic and modular solutions including their distilled versions.}
    \label{fig:latency_of_mono_vs_modu_with_distilled}
\end{figure}

% Speed experiment.
The spanish product review test set of the amazon review dataset is used to compare the latency of the two original monolithic and modular solutions and the two distilled versions of the solution. In this experiment we measure the time it takes for each model to process the 30000 test data points. The results are shown in fig. \ref{fig:latency_of_mono_vs_modu_with_distilled}. The experiments are conducted on a machine with Intel(R) Xeon(R) CPU @ 2.20GHz, 13298580 kB of RAM and a Tesla P100 16280MiB GPU.

% Discussion of accuracy and latency results.
% Discuss the observations. This will refer to the graphical experiment results.
% Put the observations in perspective with the concerns that we discussed above. Discuss the concerns we discussed above with the concrete example.
% Todo: maybe improve the writing style here.
The results in fig. \ref{fig:latency_of_mono_vs_modu_with_distilled} shows that the original modular solution is much slower compared to the original monolithic solution. This performance drop is mainly due to the translation model used in the first stage of the modular solution. However, the solutions distilled from each of the original solutions are much faster as we can expect. These results suggest that developing modular solutions and distilling them into smaller models can result in efficient solutions with minor accuracy trade-offs. The additional benefit of this approach is that the modular solution development provides a number of engineering advantages as discussed in the introduction section.

%%% Satellite
\subsubsection{Satellite Image Classification}

\begin{figure}[ht]
    \centering
    \includegraphics[width=0.46\textwidth]{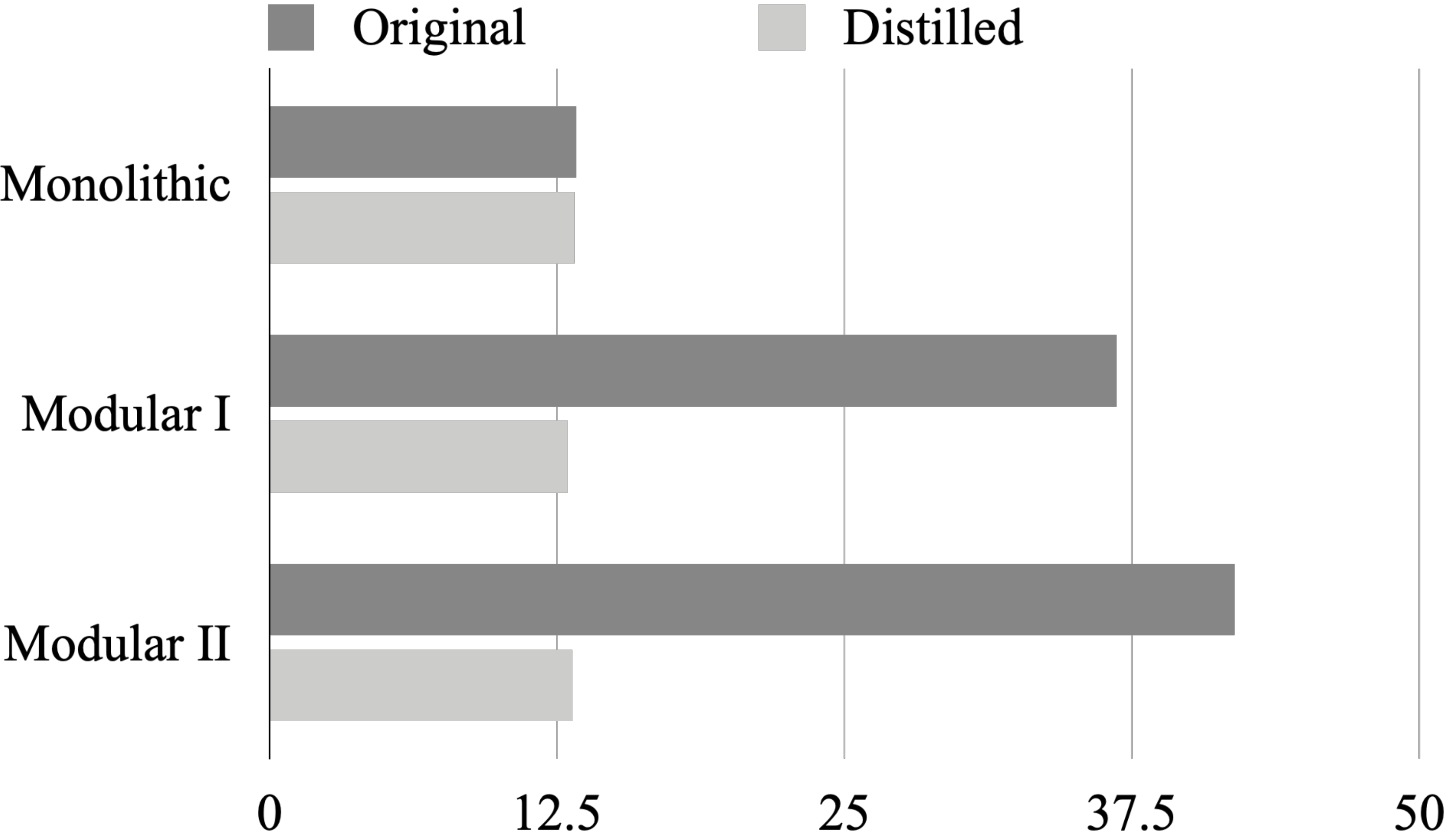}
    \caption{Cloudy satellite image classification latency (lower the better) of monolithic and modular solutions with the respective distilled versions.}
    \label{fig:mono_modu.i_modu.ii_distilled_sat_latency}
\end{figure}

% Latency experiment.
To compare the latency of the monolithic and modular solutions, the time that each solution takes to process 56700 data points is measured. The same is done with the distilled solutions. The latency measurements for each solution is taken on a machine with Intel(R) Xeon(R) CPU @ 2.20GHz, 13298580 kB of RAM and a Tesla P100 16280MiB GPU. The results are shown in fig. \ref{fig:mono_modu.i_modu.ii_distilled_sat_latency}.

The latency results in fig. \ref{fig:mono_modu.i_modu.ii_distilled_sat_latency} shows that the non-distilled monolithic solution took 63.83\% less time compared to the fastest modular solution to process the data load. However, distilled versions have been able to address this issue by achieving low latencies comparable to the monolithic version, still having better accuracy than the monolithic version in the case of modular (II) solution.

%%%%%%%%%%%%%%%%%%%%%%%%%%%%%%%%%%%%%%%%%%%%%%%%%%%%%%%
%%%%%%%%%%%%%%%%%%%%%%%%%%%%%%%%%%%%%%%%%%%%%%%%%%%%%%%
%%%%%%%%%%%%%%%%%%%%%%%%%%%%%%%%%%%%%%%%%%%%%%%%%%%%%%%
%%%%%%%%%%%%%%%%%%%%%%%%%%%%%%%%%%%%%%%%%%%%%%%%%%%%%%%
%%%%%%%%%%%%%%%%%%%%%%%%%%%%%%%%%%%%%%%%%%%%%%%%%%%%%%%
%%%%%%%%%%%%%%%%%%%%%%%%%%%%%%%%%%%%%%%%%%%%%%%%%%%%%%%
%%%%%%%%%%%%%%%%%%%%%%%%%%%%%%%%%%%%%%%%%%%%%%%%%%%%%%%

\subsection{Reusability}

% NIR experiment.
In this section we are testing whether an ML module that was used in one problem can be used in another problem. This is different from general transfer learning where base layers of a larger model are transferred to a similar task. Here we are reusing a semantically meaningful ML module in a different problem.

\begin{figure}[ht]
    \centering
    \includegraphics[width=0.46\textwidth]{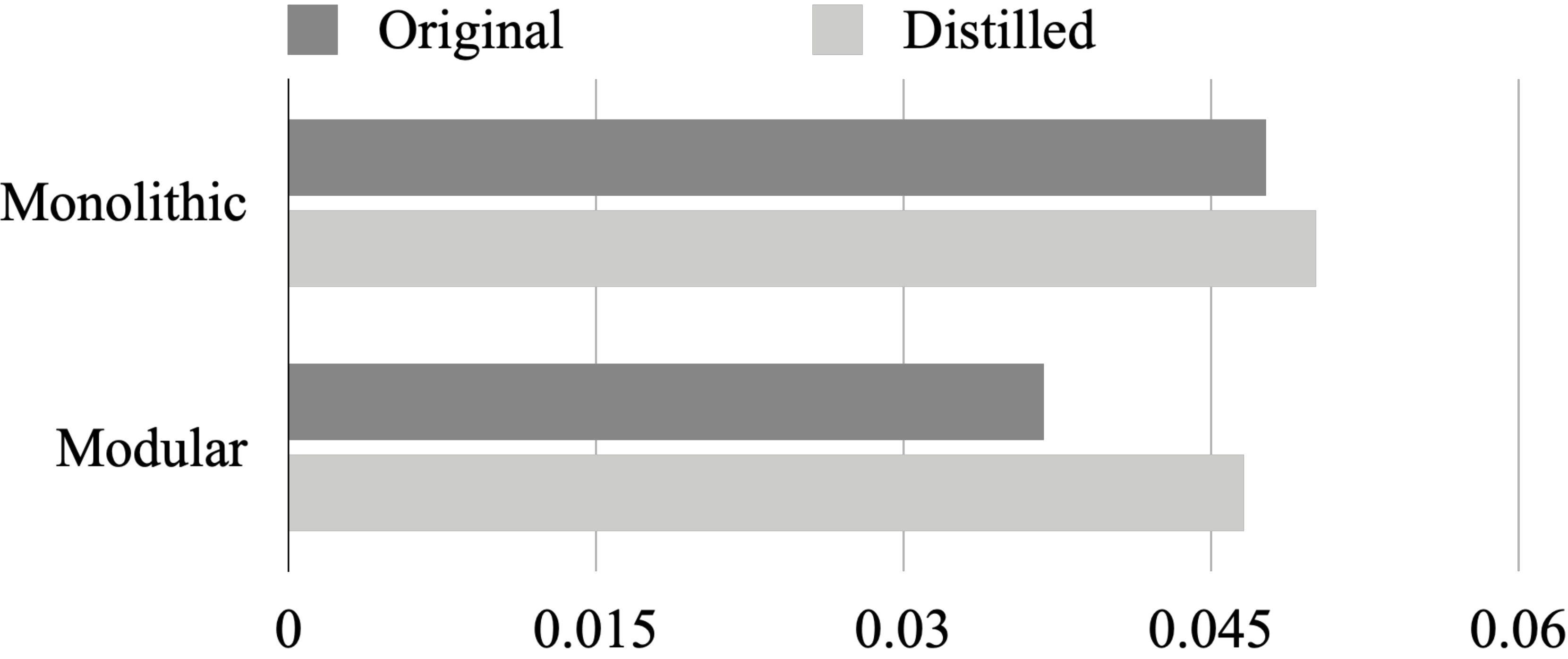}
    \caption{Mean squared error (lower the better) of monolithic and modular NIR prediction solutions including their distilled versions.}
    \label{fig:chart_mono_modu_distilled_sat_nir_mse}
\end{figure}

\begin{figure}[ht]
    \centering
    \includegraphics[width=0.46\textwidth]{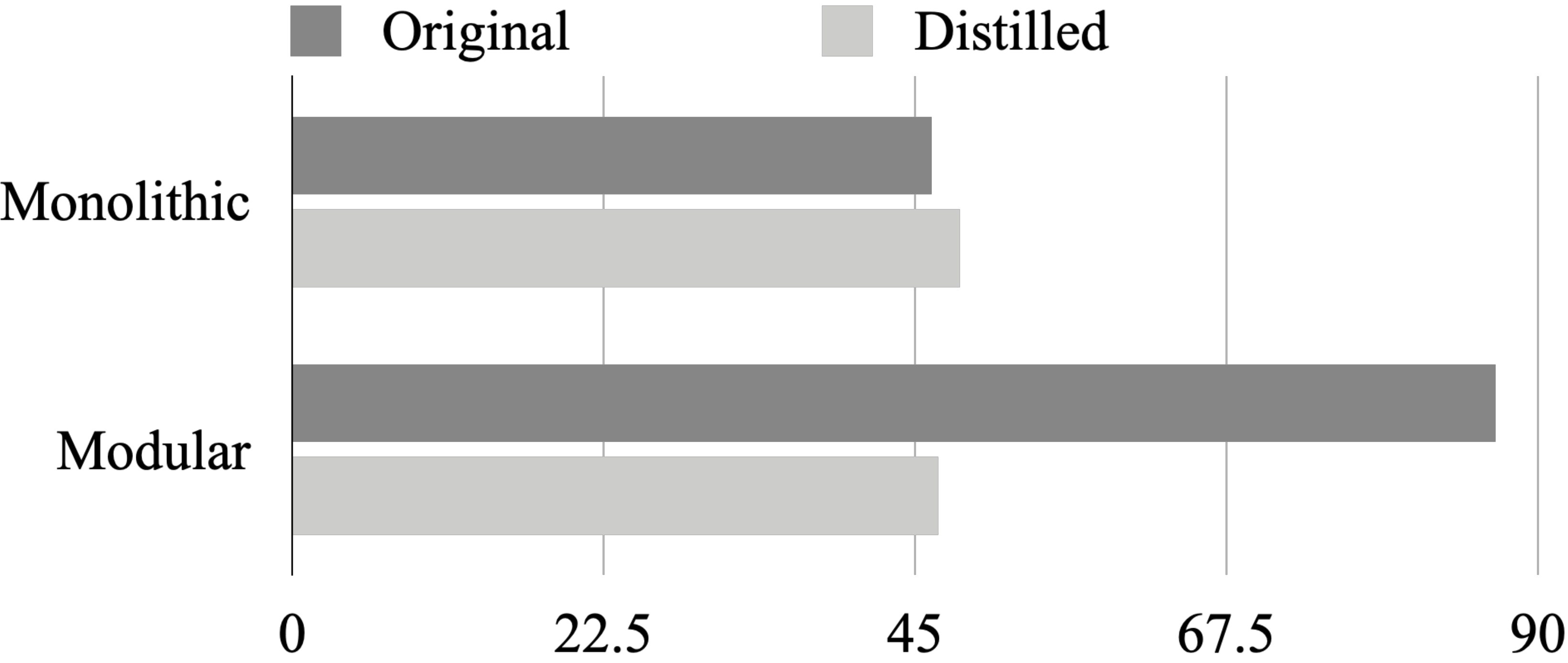}
    \caption{Latency (lower the better) of monolithic and modular NIR prediction solutions with the respective distilled versions.}
    \label{fig:chart_mono_modu_distilled_sat_nir_latency}
\end{figure}

After the classification comparison, the cloud removal module is reused in a different task to evaluate the reusability of the module. In this task the RGB bands of the cloudy EuroSAT dataset are used to predict the NIR band for the image. Two monolithic and modular solutions are implemented for this task. The monolithic solution is trained end to end on the cloudy RGB images. Monolithic model uses an encoder decoder network architecture similar to the one used in the cloud removal module but with a single output channel. The modular version predicts the NIR band in two stages. The first stage removes the clouds from training data by reusing the cloud removal module that was trained during the previous classification task. The second stage uses the output of the cloud removal module to predict the NIR band. Model used in the second stage uses the same network architecture used in the monolithic version. Two distilled models are created in this case as well using the monolithic and modular versions of the solution. 
Both modular and monolithic versions of the solutions are trained using the NIR prediction training data split that was explained before. The distillation was performed following the same process used for distilling the monolithic and modular (I) classification models. However, in this case the distilled models are pre trained with the NIR prediction dataset before the distillation. After the training and distillation steps, test split of the NIR prediction dataset is used to measure the accuracy of each solution. The results of this experiment are shown in fig. \ref{fig:chart_mono_modu_distilled_sat_nir_mse}. The latency of each solution is measured for processing 56700 data points using each version of the solution on a machine with Intel(R) Xeon(R) CPU @ 2.20GHz, 13298580 kB of RAM and a Tesla P100 16280MiB GPU. The latency results are shown in fig. \ref{fig:chart_mono_modu_distilled_sat_nir_latency}.

% Results discussion.
The results in fig. \ref{fig:chart_mono_modu_distilled_sat_nir_mse} and  \ref{fig:chart_mono_modu_distilled_sat_nir_latency} resembles the pattern we observed in the classification case with monolithic and modular (I) solutions. The modular solution has higher performance in terms of accuracy/error. However, it has higher latency compared to the monolithic version as one would expect. The solution distilled from the modular model has lower error compared to the both original monolithic solution and the solution distilled from the monolithic solution while having comparable latency values.

% Maybe it is interesting to mention that the modular versions had smoother learning curves.

%%%%%%%%%%%%%%%%%%%%%%%%%%%%%%%%%%%%%%%%%%%%%%%%%%%%%%%
%%%%%%%%%%%%%%%%%%%%%%%%%%%%%%%%%%%%%%%%%%%%%%%%%%%%%%%
%%%%%%%%%%%%%%%%%%%%%%%%%%%%%%%%%%%%%%%%%%%%%%%%%%%%%%%
%%%%%%%%%%%%%%%%%%%%%%%%%%%%%%%%%%%%%%%%%%%%%%%%%%%%%%%
%%%%%%%%%%%%%%%%%%%%%%%%%%%%%%%%%%%%%%%%%%%%%%%%%%%%%%%
%%%%%%%%%%%%%%%%%%%%%%%%%%%%%%%%%%%%%%%%%%%%%%%%%%%%%%%
%%%%%%%%%%%%%%%%%%%%%%%%%%%%%%%%%%%%%%%%%%%%%%%%%%%%%%%

\subsection{Maintainability}

In this section, we empirically highlight the trade-offs of the monolithic and the modular solutions with respect to maintainability as the requirements change. We will be considering a case where the requirements of the model change to handle noisy images in the cloudy satellite image classification problem.

\begin{figure}[ht]
    \centering
    \includegraphics[width=0.48\textwidth]{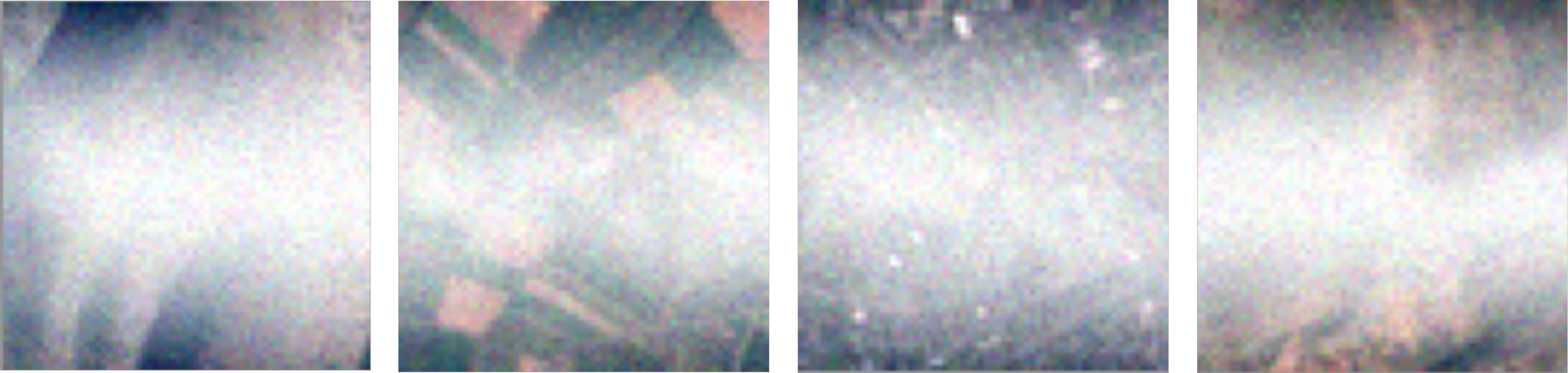}
    \caption{Cloudy EuroSAT data points after adding gaussian noise.}
    \label{fig:data_eurosat_cloud_added_noisy}
\end{figure}

\begin{figure}[ht]
    \centering
    \includegraphics[width=0.46\textwidth]{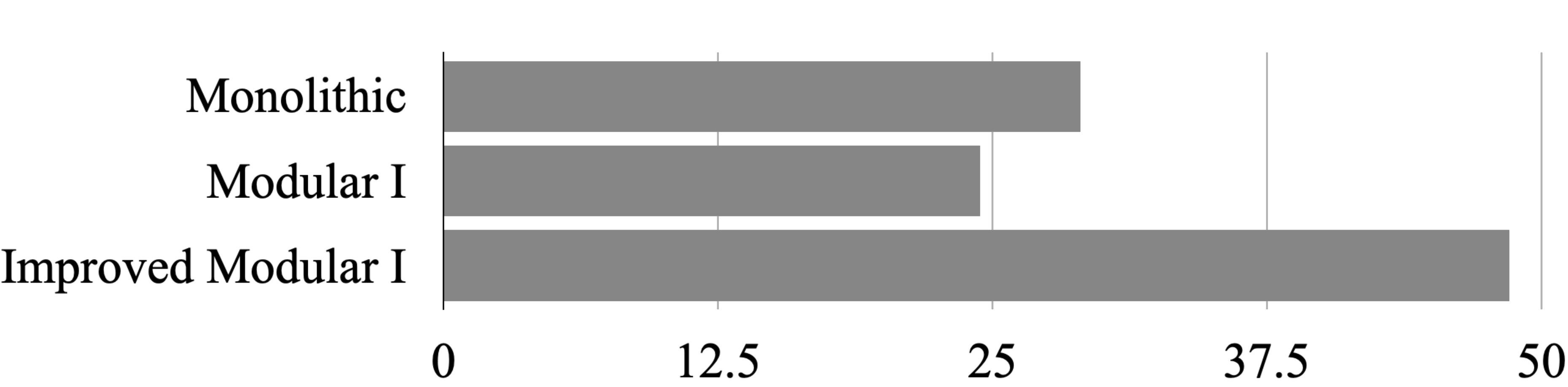}
    \caption{Classification accuracy for noisy satellite images with cloud cover.}
    \label{fig:chart_mono_modu.i_modu.i.improved_noisy_cloudy_accuracy}
\end{figure}

In this experiment, we modify the classification dataset that we used before by adding gaussian noise to the RGB channels to represent noisy satellite images with cloud cover. Fig. \ref{fig:data_eurosat_cloud_added_noisy} shows the images after adding noise. 
Then the noisy data is fed to the previously trained monolithic classification model and the modular (I) classification model and the classification accuracy is measured using a held out test set. Additionally, we create an improved modular solution by updating the cloud removal module of the modular (I) solution. The cloud removal module is updated by training it with unlabeled noisy cloud images. This improvement makes the cloud removal module robust to noise. In the improved modular solution, the classification module remains to be the same module used in the original modular (I) solution. The accuracy of the improved modular solution is measured with the same noisy held out test data. Accuracy of each model is shown in fig. \ref{fig:chart_mono_modu.i_modu.i.improved_noisy_cloudy_accuracy}.
As shown in the results in fig. \ref{fig:chart_mono_modu.i_modu.i.improved_noisy_cloudy_accuracy}, both monolithic and modular (I) solutions significantly drop in accuracy when fed with noisy images. However, the modular (I) solution has the capacity to replace modules with improved ones. By replacing the cloud removal module with an updated noise robust cloud removal module, the improved modular solution could achieve around 2 times higher accuracy compared to the original  modular (I) solution. It should be noted that this accuracy gain was achieved without making any changes to the classification module and by only using noisy unlabeled data. The monolithic solution can not be improved this way using unlabeled data. Improving the monolithic solution needs labeled data that are usually labor intensive to produce.

% Talk about the ability to replace parts and achieve different characteristics. Talk about debuggability.

% Improve the cloud removal module and show that it increases classif accuracy and also, nir accuracy.
% Debug an image and show it is easier to do in the modular solution.

%%%%%%%%%%%%%%%%%%%%%%%%%%%%%%%%%%%%%%%%%%%%%%%%%%%%%%%
%%%%%%%%%%%%%%%%%%%%%%%%%%%%%%%%%%%%%%%%%%%%%%%%%%%%%%%
%%%%%%%%%%%%%%%%%%%%%%%%%%%%%%%%%%%%%%%%%%%%%%%%%%%%%%%
%%%%%%%%%%%%%%%%%%%%%%%%%%%%%%%%%%%%%%%%%%%%%%%%%%%%%%%
%%%%%%%%%%%%%%%%%%%%%%%%%%%%%%%%%%%%%%%%%%%%%%%%%%%%%%%
%%%%%%%%%%%%%%%%%%%%%%%%%%%%%%%%%%%%%%%%%%%%%%%%%%%%%%%
%%%%%%%%%%%%%%%%%%%%%%%%%%%%%%%%%%%%%%%%%%%%%%%%%%%%%%%

\section{Open Challenges}

Modularity in ML has a number of advantages. We discussed them in detail qualitatively and quantitatively. However, modular ML has several main open challenges when it comes to developing effective modular ML models. 

Breaking down the original problem into a set of subproblems is not always feasible. Some problems do not have semantically meaningful subproblems. In these cases we have to consider the problem as an atomic unit and utilize end to end learning methods. After that the trained model has the potential to be used in a future Modular ML model.

In situations where new models have to be trained to solve subproblems, finding datasets for them can sometimes be challenging. This issue can be mitigated by performing the problem decomposition to match data that is available. In this case, the trade-offs of different problem decompositions should be further studied. Further, synthetic data can be helpful to fill some of the data gaps.

In some cases, module compositions may not perform in synergy as expected. Sometimes even if the individual models perform well, when they are composed together to solve a larger problem, the performance can be unexpectedly low. This can happen due to various incompatibilities among submodules. Studies on ML adversarial attacks may be able to shed some light in this regard. Further understanding the reasons for such failures and methods is important to make modular ML applicable in a wider array of problems.

Lack of feature rich repositories to publish and search ML models and datasets is another challenge to using modular ML in practice. Existing systems for publishing and searching ML models do not provide sufficient metadata and semantically meaningful search capabilities to look for models and datasets that can satisfy specific requirements. More advanced ML models and data repositories with semantic search capabilities and metadata support~\cite{DBLP:conf/semco/MenikR21} should be developed to make modular ML practical.

Addressing these challenges through future work is key to reaping the benefits of modular machine learning.

%%%%%%%%%%%%%%%%%%%%%%%%%%%%%%%%%%%%%%%%%%%%%%%%%%%%%%%
%%%%%%%%%%%%%%%%%%%%%%%%%%%%%%%%%%%%%%%%%%%%%%%%%%%%%%%
%%%%%%%%%%%%%%%%%%%%%%%%%%%%%%%%%%%%%%%%%%%%%%%%%%%%%%%
%%%%%%%%%%%%%%%%%%%%%%%%%%%%%%%%%%%%%%%%%%%%%%%%%%%%%%%
%%%%%%%%%%%%%%%%%%%%%%%%%%%%%%%%%%%%%%%%%%%%%%%%%%%%%%%
%%%%%%%%%%%%%%%%%%%%%%%%%%%%%%%%%%%%%%%%%%%%%%%%%%%%%%%
%%%%%%%%%%%%%%%%%%%%%%%%%%%%%%%%%%%%%%%%%%%%%%%%%%%%%%%

\section{Related Work}

% Talk about approaches that tried to break trained models. But did not go far.

% There are existing approaches that encourage modularity in machine learning systems.
There are several existing lines of work that are related to modularity in machine learning.
% Breaking monolithic models.
There have been attempts to break pre-trained neural networks into a set of modules based on the learned weights. Ultimate goal of this is to find semantic modules within a learned neural network. Csordas et. al.~\cite{DBLP:conf/iclr/CsordasSS21} have attempted to do this by learning weight masks to identify subnetworks for target tasks. They were able to find some specialized subnetworks in trained neural networks with some other interesting insights. However, to the best of our knowledge, these works have so far not been able to find strong semantically meaningful modules within learned monolithic neural networks that can be reused in other contexts.
% Ensemble learning.
% Cascading modules.
Ensemble learning methods \cite{sagi2018ensemble} has a sense of modularity. An ensemble model is not a single monolithic unit. Ensemble methods like bagging, stacking and boosting combine several machine learning models to create a more robust single solution. In these methods each model in the ensemble will be specialized in one aspect of the whole problem. These specialized parts together create a better final solution to the overall machine learning problem. However, the modules in ensemble models are usually not reusable in other systems since they are specialized to a given specific ensemble that solves one problem.
% Modular deep learning. Routing networks (maybe remove this topic).
Another line of work that attempts to incorporate modularity in deep learning is routing networks \cite{cases2019recursive}. Modular question answering approach proposed by Jacobs \cite{andreas2016learning} is another example for similar work. This line of work mainly tries to learn a set of modules and a controller to compose these modules to solve problems. This learning process is usually done in an end to end fashion. Even though these approaches have shown good performance in the problems that the model was trained to, the modules have not shown to be much useful outside of the problem in concern. Therefore, routing networks, in their current state, are not yet capable of addressing the problems that we discussed earlier.
% Transfer learning
Transfer learning techniques \cite{zhuang2020comprehensive} allow machine learning model developers to train customized solutions with less amount of training data by fine tuning an already existing machine learning model that is usually pre-trained with a large dataset. Transfer learning has shown to be very effective when an already existing model is similar to the target task. There are few downsides to transfer learning approaches. First, the model developers usually have to be knowledgeable about the pre-existing model internals in order to be able to modify it to match the target machine learning problem. The modified model has to be retrained with new training data that better represent the new machine learning problem. This is different from traditional software engineering modules that attempt to expose a clean interface to the users by hiding the module internals. When it comes to making machine learning technologies more accessible to a wider audience, transfer learning techniques should be further improved to enable a more ready to use modularity that minimizes the chances of developers having to work with the internals of complex network architecture.

\section{Summary}

In this work, while acknowledging the immense potential and positive impact of current deep learning technologies, we discussed the challenges and limitations of widespread monolithic deep learning technologies with respect to systems engineering concerns especially when it comes to wider adoption of these technologies in diverse organizations. We pointed out semantic modularity in machine learning as an interesting avenue to address a number of problems in this regard. Next, as a first step, we used three example problems to explore the benefits and trade-offs between developing machine learning solutions in a monolithic way and developing them in a multi-stage modular way. The experiments showed how modular solutions can reuse existing pretrained models and exploit more data to achieve higher accuracy and overcome data limitations in ways that monolithic solutions do not permit. Further, we used blackbox knowledge distillation to overcome the performance challenges that modular solutions can have and showed the impact of accuracy and latency in comparison to original monolithic and modular solutions. The experimental results in this work suggest that it is very interesting to further investigate the potential of multi stage modular machine learning solution development in contrast to widespread monolithic end to end machine learning solution development.

%%%%%%%%%%%%%%%%%%%%%%%%%%%%%%%%%%%%%%%%%%%%%%%%%%%
%%%%%%%%%%%%%%%%%%%%%%%%%%%%%%%%%%%%%%%%%%%%%%%%%%%
%% End of content.
%%%%%%%%%%%%%%%%%%%%%%%%%%%%%%%%%%%%%%%%%%%%%%%%%%%
%%%%%%%%%%%%%%%%%%%%%%%%%%%%%%%%%%%%%%%%%%%%%%%%%%%

\bibliography{references}
\bibliographystyle{mlsys2023}

%%%%%%%%%%%%%%%%%%%%%%%%%%%%%%%%%%%%%%%%%%%%%%%%%%%%%%%%%%%%%%%%%%%%%%%%%%%%%%%
%%%%%%%%%%%%%%%%%%%%%%%%%%%%%%%%%%%%%%%%%%%%%%%%%%%%%%%%%%%%%%%%%%%%%%%%%%%%%%%
% SUPPLEMENTAL CONTENT AS APPENDIX AFTER REFERENCES
%%%%%%%%%%%%%%%%%%%%%%%%%%%%%%%%%%%%%%%%%%%%%%%%%%%%%%%%%%%%%%%%%%%%%%%%%%%%%%%
%%%%%%%%%%%%%%%%%%%%%%%%%%%%%%%%%%%%%%%%%%%%%%%%%%%%%%%%%%%%%%%%%%%%%%%%%%%%%%%
% \appendix
% \section{Please add supplemental material as appendix here}
% %
% Put anything that you might normally include after the references as an appendix here, {\it not in a separate supplementary file}. Upload your final camera-ready as a single pdf, including all appendices.

%%%%%%%%%%%%%%%%%%%%%%%%%%%%%%%%%%%%%%%%%%%%%%%%%%%%%%%%%%%%%%%%%%%%%%%%%%%%%%%
%%%%%%%%%%%%%%%%%%%%%%%%%%%%%%%%%%%%%%%%%%%%%%%%%%%%%%%%%%%%%%%%%%%%%%%%%%%%%%%

\end{document}